\documentclass{article}
\pdfoutput=1

\usepackage{arxiv}

\usepackage[utf8]{inputenc} 
\usepackage[T1]{fontenc}    
\usepackage{hyperref}       
\usepackage{url}            
\usepackage{booktabs}       
\usepackage{amsfonts}       
\usepackage{nicefrac}       
\usepackage{microtype}      
\usepackage{graphicx} 
\usepackage{amsmath, amsthm, amssymb}
\usepackage{tabularx}
\usepackage{color}
\usepackage{natbib}
\usepackage{algorithm}
\usepackage{algorithmic}

\newtheorem{proposition}{Proposition}  
\newtheorem{definition}{Definition}
\newtheorem{theorem}{Theorem}
\newtheorem{remark}{Remark}
\newtheorem{lemma}{Lemma}

\DeclareMathOperator\chol{chol}

\def\U{{\mathbf{U}}}
\def\u{{\mathbf{u}}}
\def\v{{\mathbf{v}}}
\def\t{{\mathbf{t}}}
\def\s{{\mathbf{s}}}
\def\x{{\mathbf{x}}}
\def\y{{\mathbf{y}}}
\def\z{{\mathbf{z}}}
\def\tto{\bar{\mathbf{t}}}

\def\no{\bar n}

\def\xo{\bar{\mathbf{x}}}
\def\yo{\bar{\mathbf{y}}}
\def\GP{{\mathcal{GP}}}
\def\WGP{{\mathcal{WGP}}}
\def\O{{\mathcal{O}}}
\def\R{{\mathbb{R}}}

\def\N{{\mathcal{N}}}

\def\X{{\mathcal{X}}}
\def\Y{{\mathcal{Y}}}
\def\I{{\mathcal{I}}}
\def\T{{\mathcal{T}}}

\def\E{{\mathbb{E}}}
\def\L{{\mathcal{L}}}

\def\xs{\lVert \x \rVert_1}
\def\ys{\lVert \y \rVert_1}

\def\xn{\lVert \x \rVert_2}
\def\yn{\lVert \y \rVert_2}

\def\xon{\lVert \xo \rVert_2}
\def\yon{\lVert \yo \rVert_2}

\def\lv{\left\vert} 
\def\rv{\right\vert} 
\def\lV{\left\Vert} 
\def\rV{\right\Vert} 
\graphicspath{ {images/} }

\title{Transport Gaussian Processes for Regression}

\author{Gonzalo Rios\\University of Chile}

\begin{document}
\maketitle


\begin{abstract}
	Gaussian process (GP) priors are non-parametric generative models with appealing modelling properties for Bayesian inference: they can model non-linear relationships through noisy observations, have closed-form expressions for training and inference, and are governed by interpretable hyperparameters. However, GP models rely on Gaussianity, an assumption that does not hold in several real-world scenarios, e.g., when observations are bounded or have extreme-value dependencies, a natural phenomenon in physics, finance and social sciences. Although beyond-Gaussian stochastic processes have caught the attention of the GP community, a principled definition and rigorous treatment is still lacking. In this regard, we propose a methodology to construct stochastic processes, which include GPs, warped GPs, Student-t processes and several others under a single unified approach. We also provide formulas and algorithms for training and inference of the proposed models in the regression problem. Our approach is inspired by layers-based models, where each proposed layer changes a specific property over the generated stochastic process. That, in turn, allows us to push-forward a standard Gaussian white noise prior towards other more expressive stochastic processes, for which marginals and copulas need not be Gaussian, while retaining the appealing properties of GPs. We validate the proposed model through experiments with real-world data. 
\end{abstract}

\section{Introduction}

In machine learning, the Bayesian approach is distinguished since it assumes a priori distribution over the possible models. As we obtain data (a.k.a observations), the Bayes rule allows us to trace the most plausible models that explain the data. For regression tasks, the Bayesian approach allows us to consider the Gaussian process as a prior over functions, which have analytical expressions and algorithms for training and inference. The main reason for its widespread use is the same as its limitation. Gaussianity assumption generates simplicity in its formulation, but in turn, causes a limited expressiveness (e.g. it fails to model a bounded domain on data). Some authors have defined other models much more expressive than GPs \cite{wilson2011gaussian}, providing methods and approximation techniques, since their exact inference is intractable \cite{krauth2016autogp}. Our primary motivation is to extend the Gaussian process methods to other stochastic processes that are more accurate in their assumptions concerning the modelled data, maintaining the elegance and interpretability of its elements.

In the literature we can find some works that address this problem, obtaining exciting and practical results. One of the first advances in this topic was the model known as the warped Gaussian process (WGP) \cite{snelson2004warped}, which is based on applying a non-linear parametric transformation to the data, so that the transformed data can be modelled with a GP in a better way than the original data. Following this idea, the model known as Bayesian warped Gaussian process (BWGP) \cite{lazaro2012bayesian} is introduced, in which a non-parametric version of the non-linear transformation is proposed. Furthermore, the interpretation is reversed: instead of transforming the data, the Gaussian process is, i.e. the result is a push-forward measure. However, analytical inference in the BWGP model is intractable, so the author gives a variational lower bound for training, and an integral formula for the one-dimensional predictive marginal, with explicit expressions for their mean and variance only.

Another related model is the deep Gaussian process (DGP) \cite{damianou2013deep}, which has been proposed primarily as a hierarchical extension of the Bayesian Gaussian process latent variable model (GP-LVM) \cite{titsias2010bayesian}, which, in turn, is a deep belief network based on Gaussian process mappings, and it focuses initially on unsupervised problems (unobserved hidden inputs) about discovering structure in high-dimensional data \cite{lawrence2004gaussian, li2016review, damianou2016variational}. However, by replacing the latent inputs with observed input, a one-hidden-layer model coincides with BWGP, so DGP for regression is also a generalisation of BWGP \cite{damianou2015deep}.

DGP is one GP feeding another GP, so it is a flexible model that can capture highly-nonlinear functions for complex data sets. However, the network structure of a DGP makes inference computationally expensive; even the inner layers has an identified pathology \cite{duvenaud2014avoiding}. To use DGP in regression scenarios, some authors propose making inference via variational approximations \cite{bui2016deep, salimbeni2017doubly} or using sequential sampling approach \cite{wang2016sequential}. Finally, DGP loses its interpretability, so, like other deep models, it is difficult to understand the properties of each layer and component.

A different related model is the Student-t process \cite{shah2014student} (SP), an extension of the GP with the appealing closed-form formulas for training and prediction. It is strictly more flexible due to heavier tails, stability against outliers and stronger dependencies structures. In practice, it has better performance than GPs on Bayesian optimisation \cite{shahriari2015taking} and state-space model regression \cite{solin2015state}. However, this model is treated entirely different from the previous models, and to date we do not know of any work that relates them in any way.

In this work, we introduce a model based on finite-dimensional maps to generate, from a reference Gaussian process noise, more expressive stochastic processes. The proposed approach can model non-Gaussian copula and marginals, beyond the known warped Gaussian process \cite{snelson2004warped, rios2018learning, riostobar2019cwgp} and Student-t process \cite{shah2014student}, but including all of them from a unifying point of view. The main idea is to construct stochastic processes, composed of different layers, following the same guidelines as deep architectures, but where each layer has an interpretation defining a feature of the process. We decompose the stochastic process on their marginals, correlation and copula, each of them isolated and characterised by ad-hoc transports. Our main contribution is to understand the well-order in compositions, to derive general analytic expressions for their posterior distributions and likelihoods functions, and to develop practical methods for the inference and training of our model, given data.

The remainder of this work is organised as follows. In Section \ref{sec:backgroundtgp}, we introduce the notation and necessary mathematical background to develop our work. Our main definition is in Section \ref{sec:deftgp}, where we propose the transport process (TP) and the inference approach. On Section \ref{sec:marginaltransport}, we study the marginal transport that isolates all properties over the univariate marginals of the TP. Similarly, in Section \ref{sec:covariancetransport}, we develop the covariance transport, that determines the correlation over the TP. Finally, the main contribution is in Section \ref{sec:familytgp}, where we introduce the radial transports, that allow us to define the dependency structure (a.k.a copula) over the TP. On Section \ref{sec:computationtgp}, we deepen in details over the computational and algorithmic implementation, and on Section  \ref{sec:experimentstgp} we validate our approach with real-world data, to finish with conclusions in Section \ref{sec:conclusionstgp}.

\section{Background}
\label{sec:backgroundtgp}

Given $N \in \mathbb{N}$ observations $(\t,\x)=\{(t_i, x_i)\}_{i=1}^N$ where $t_i \in \mathcal{T}\subseteq \R^T$, $T \in \mathbb{N}$ and $x_i \in \X \subseteq \R$ for $i=1,\ldots,n$ the regression problem aims to find the \emph{best} predictor $f:\mathcal{T} \rightarrow \mathcal{X}$, such that $f(t_i)$ is \emph{close} to $x_i$, where the terms \emph{best} and \emph{close} are given by the chosen criterion of optimality. In several fields, such as finance, physics and engineering, we can find settings where the observations are indexed by time or space and convey some hidden dependence structure that we aim to discover. A Bayesian non-parametric solution to this regression problem are the Gaussian processes \cite{rasmussen06}, also know as kriging \cite{stein2012interpolation,cressie1990origins}.

\begin{definition}
	A stochastic process $f=\{x_{t}\}_{t\in \mathcal{T}}$ is a Gaussian process (GP) with mean function $m(\cdot)$ and covariance kernel $k(\cdot,\cdot)$, denoted by $f \sim \mathcal{GP}\left( m,k\right)$, if, for any finite collection of points in their domain  $\t=[t_1, \ldots, t_n]^\top \in \mathcal{T}^n$, the distribution $\eta_{\t}$ of the vector\footnote{By abuse of notation, we identify the random vector $f(\t)$ as $\x$, which denote the observations on $\t$.} $\x:=f(\t) = [x_{t_1}, \ldots, x_{t_n}]^\top\in\mathcal{X}^n$ follows a multivariate Gaussian distribution with mean vector $\mu_{\x} = [m(t_1), \ldots, m(t_n)]^\top$ and covariance matrix $[\Sigma_{\x\x}]_{ij} = k(t_i,t_j)$, i.e. $\eta_{\t} = \N_{n}(\mu_{\x}, \Sigma_{\x\x})$.
\end{definition}

For a distribution $\eta_{\t}$ that depends on parameters $\theta$\footnote{As long as there is no ambiguity in inputs points $\t$ and parameters $\theta$, we denote the evaluated process $f(\t)$ as $\x$, its mean $m(\t)$ as $\mu_{\x}$ and its covariance $k(\t,\t)$ as $\Sigma_{\x\x}$, without referencing $\theta$. For a second collection of input points $\tto$ the notation is analogue: the process evaluation is $\xo=f(\tto)$, the mean is $\mu_{\xo}=m(\tto)$ and the cross-covariance between $\x$ and $\xo$ is $\Sigma_{\x\xo}=k(\t,\tto)$.}, we denote the evaluation of their density function at $\x$ as $\eta_{\t}(\x|\theta)$. Thus, given observations $(\t,\x)$, learning is equivalent to inferring $m(\cdot)$ and $k(\cdot,\cdot)$, finitely-parameterised by $\theta=(\theta_k,\theta_m) \in \R^p$. This is achieved by minimising the negative logarithm of the marginal likelihood \footnote{In practice, we choose a parametrisation of $m(\cdot)$ and $k(\cdot,\cdot)$, so the NLL is continuous and derivable w.r.t parameters $\theta$. However, the main difficulty is that the resulting functional is non-linear and populated with multiple local optima.} (NLL), given by 
\begin{align}
	 -\log\eta_{\t}(\x|\theta) = \frac{n}{2}\log(2\pi) + \frac{1}{2}\left(\x-\mu_{\x} \right)^{\top}\Sigma_{\x \x}^{-1}\left(\x-\mu_{\x} \right) + \frac{1}{2} \log \left|\Sigma_{\x \x}\right|.
\end{align}

Performing prediction on new inputs $\tto$ rests on inference $\xo$ given observations $\x$, given by the posterior distribution of which is also Gaussian and has distribution $\eta_{\tto|\t} = \N\left(\mu_{\xo|\x} , \Sigma_{\xo|\x}\right)$ where $\mu_{\xo|\x} = \mu_{\xo}+\Sigma _{\xo\x}\Sigma _{\x\x}^{-1}\left(\x -\mu_{\x}\right)$ and $\Sigma_{\xo|\x} = \Sigma _{\xo\xo}-\Sigma _{\xo\x}\Sigma _{\x\x}^{-1}\Sigma _{\x\xo}$ are referred to as the conditional mean and variance respectively.

\subsection{The Gaussian assumption}

Both the meaningfulness and the limitations of the GP approach stem from a common underlying assumption: Gaussian data. For instance, under the presence of strictly-positive observations, e.g. prices of a currency or the streamflow of a river, assuming Gaussianity is a mistake, since the Gaussian distribution is supported on the entire real line. A standard practice in this case is to transform the observed data $\y \in \Y^N$ via a non-linear differentiable bijection $\varphi:\Y \rightarrow \X$ such that $\x=\Phi(\y)=[\varphi(y_1),...,\varphi(y_N)]^\top$ is ``more Gaussian'' and thus can be modeled as a GP. A common choice for such a map is $\varphi(y) = \log(y)$, where the implicit assumption is that the observed process has log-normal marginals and, in particular, positive values. This generative model, named warped Gaussian process (WGP) \cite{snelson2004warped}, has a closed-form expression for the density of $\y$, thanks to the change of variables formula \cite{hogg1995introduction}, enunciated below:

\begin{theorem}
	\label{thm:CoV}
	Let $\x \in \X \subseteq \R^{n}$ be a random vector with a probability density function given by $p_{\x}\left(\x\right)$, and let $\y\in \Y\subseteq \R^n$ be a random vector such that $\Phi \left(\y\right) = \mathbf{x}$, where the function  $\Phi :\Y\rightarrow \X$ is bijective of class $\mathcal{C}^{1}$ and $\left\vert \nabla \Phi\left( \y\right) \right\vert >0$, $\forall \y\in \Y$. Then, the probability density function $p_\y(\cdot)$ induced in $\mathcal{Y}$ is given by $p_{\y}\left( \y\right) =p_{\x}\left( \Phi \left( \y\right)\right) \left\vert \nabla \Phi \left( \y\right) \right\vert$, where $\nabla \Phi \left( \cdot\right)$ denotes the Jacobian of $\Phi \left( \cdot\right)$, and $|\cdot|$ denotes the determinant operator.
\end{theorem}

The warped GP is a well-defined stochastic process since the transformation $\Phi$ (the \emph{transport map}) is diagonal (i.e. defined in a coordinate-wise manner $\Phi(\y)_{i} = \varphi(x_i)$), so the induced distributions satisfy the conditions of the \emph{Kolmogorov’s consistency theorem} \cite{tao2011introduction}. On section \ref{sec:deftgp} we will define and study this consistency property in detail.

\subsection{The dependence structure}
\label{sec:copulatgp}

Warped GPs define non-Gaussian models with appealing mathematical properties akin to GPs, such as having closed-form expressions for inference and learning. However, they inherit an unwanted Gaussian drawback: the dependence structure in this class of processes remains purely Gaussian. To understand the implications of this issue, we need to formalise the concept of dependence and some essential related results. Let us fix some notation and conventions. 


Given a multivariate distribution $\eta$, we denote its cumulative distribution function by $F_\eta(\cdot)$. As long as there is no ambiguity, the cumulative distribution function of their $i$-th marginal distribution $\eta_{i}$ is denoted as $F_i(x) := F_{\eta_i}(x)$, as well as its right-continuous quantile function, $Q_i(u) := F_i^{-1}(u) = \inf \{x| F_i(x) \ge u\}$. If a multivariate cumulative distribution function $C$ has uniform univariate marginals, that is, $C_i(u) = \max(0, u \wedge 1)$ for $i=1,...,n$, then we say that $C$ is a \emph{copula}. The next result, known as Sklar's theorem \cite{sklar1959fonctions}, shows that any distribution has a related copula.

\begin{theorem}
	Given a multivariate distribution $\eta$, there exists a copula $C$ such that $F_{\eta}(x_{1},...,x_{n}) = C(F_{1}(x_{1}),...,F_{n}(x_{n}))$. If the $F_i$ are continuous, for $i=1,...,n$, then the copula is unique and given by $C_{\eta}(u_{1},...,u_{n}) = F_{\eta}(F^{-1}_{1}(u_{1}),...,F^{-1}_{n}(u_{n}))$.
\end{theorem}

If $\eta$ is a Gaussian distribution, its unique copula has a density determined entirely by its correlation matrix $R$, and it is given by $c_{\eta}(\u) = \det(R)^{-\frac{1}{2}} \exp \left(-\frac{1}{2}\x^{\top}[R^{-1}-I]\x\right)$, where $x_i = F_{s}^{-1}(u_i)$ with $F_{s}$ the standard normal cumulative distribution function. Note that if their coordinates are uncorrelated, then $C_\eta$ coincides with the independence copula.

\subsection{The devil is in the tails}

For Gaussian models, correlation and dependence are equivalent; however, beyond the realm of Gaussianity, this is not the case. In the general case, dependence between variables is more complex than just correlation, highlighting an extreme value theory concept: tail dependence \cite{coles2001introduction}. Some variables can be uncorrelated but can show dependence on extreme deviations, as exhibited in financial crises or natural disasters. Unfortunately, as outlined below, the Gaussian copula is not suitable for these kinds of structural dependences.

The coefficients of lower and upper tail dependence between two variables $x_{1}$ and $x_{2}$ are defined as $\lambda_{l} = \lim_{q \to 0} \mathbb{P}\left(x_{2} \leq F_{2}^{-1}(q) |  x_{1} \leq F_{1}^{-1}(q)\right)$ and $\lambda_{u} = \lim_{q \to 1} \mathbb{P}\left(x_{2} > F_{2}^{-1}(q) |  x_{1} > F_{1}^{-1}(q)\right)$ \cite{schmidt2005tail} . These coefficients provide asymptotic measures of the dependence in the tails (extreme values), which are isolates of their marginals distributions. For independent continuous r.v. we have that $\lambda_{l} = \lambda_{u} = 0$, whereas for variables with correlation $\rho=1$ we have that $\lambda_{l} = \lambda_{u} = 1$. For Gaussian distributions, however, the result is surprising: for $\rho < 1$ we have that $\lambda_{l} = \lambda_{u} = 0$.

The above result implies that Gaussian variables are \emph{asymptotically independent}, meaning that the Gaussian assumption does not allow for modelling extreme values dependence. This inability, inherited by any diagonal transformation such as $\Phi$ aforementioned, can result in misleading calculations of probabilities over extreme cases. This issue was observed mainly in the 2008 subprime crisis, where the Gaussian dependence structure is pointed out as one of the leading causes, thus evidencing that \emph{the devil is in the tails} \cite{donnelly2010devil}. Constructing stochastic processes that account for tail dependence is challenging since, in general, distributions satisfying the consistency conditions are scarce.


\section{Transport Process}
\label{sec:deftgp}

While the measure-theoretic approach to stochastic processes starts with a probability space, in machine learning the starting point is a collection of finite-dimensional distributions. The well-know \emph{Kolmogorov's consistency theorem} \cite{tao2011introduction} guarantees that a suitably \emph{consistent} collection of these distributions $\mathcal{F} = \{\eta_{t_{1},...,t_{n}}| t_{1},...,t_{n} \in \mathcal{T}, n\in \mathbb{N} \}$ will define a stochastic process $f=\{x_{t}\}_{t\in \mathcal{T}}$, with finite-dimensional laws $\mathcal{F}$. By abuse of notation, their law is denoted as $\eta$. Denoting by $F_{t_{1},...,t_{n}}(x_1,...,x_n)$ the cumulative distribution function of $\eta_{t_{1},...,t_{n}}$, the \emph{consistency} conditions over $\mathcal{F}$ are:
\begin{enumerate}
	\item Permutation condition: $F_{t_{1},...,t_{n}}\left( x_{1},...,x_{n}\right) =F_{t_{\tau \left( 1\right)},...,t_{\tau \left( n\right) }}\left( x_{\tau \left( 1\right) },...,x_{\tau\left( n\right) }\right)$ for all $t_1,...,t_{n} \in \T$, all $x_1,...,x_n \in \X$ and any $n$-permutation $\tau$.
	\item Marginalisation condition: $F_{t_{1},...,t_{n+m}}\left( x_{1},...,x_{n},+\infty,...,+\infty \right) =F_{t_{1},...,t_{n}}\left( x_{1},...,x_{n}\right)$ for all $t_1,...,t_{n+m} \in \T$ and all $x_1,...,x_n \in \X$.
\end{enumerate}

The main idea that we develop in this paper is, for a given and fixed reference stochastic process $f$, \emph{push-forwarding}\footnote{Given a measure $\eta$ and a measurable map $T$, the \emph{push-forward} of $\eta$ by $T$ is the measure defined as $[T\#\eta](\cdot) = \eta(T^{-1}(\cdot))$.} each of its finite-dimensional laws $\eta_{\t} \in \mathcal{F}$ by some measurable maps $T_{\t} \in T$\footnote{Since the set of all indexed measurable maps $T_{\t}$ contains information on all coordinates, by abuse of notation it is denoted as $T$.}, to generate a new set of finite-dimensional distributions $\mathcal{\hat F}$ and thus a stochastic process. The main difficulty of this approach is that, in general, $\mathcal{\hat F}$ can be inconsistent, in the sense that it can violate some consistency conditions; however, it is possible to choose the maps that induce a consistent set of finite-dimensional laws and therefore a stochastic process. 

The following definition is one of our main contributions as it allows us to construct non-Gaussian processes as non-parametric regression models. 

\begin{definition}
	Let $T=\{T_\t:\X^n \to \Y^n \subseteq \R^n | \t \in \T^n, n \in \mathbb{N}\}$ be a collection of measurable maps and $f=\left\{ x_{t}\right\} _{t\in \T}$ a stochastic process with law $\eta$. We say that $T$ is a $f$-transport if the push-forward finite-dimensional distributions $\mathcal{\hat F}=\{\pi_{\t}:=T_{\t}\#\eta_{\t}| \t \in \T^n, n \in \mathbb{N}\}$ are consistent and define a stochastic process $g=\left\{ y_{t}\right\} _{t\in \T}$ with law $\pi$. In this case we say that the maps $T_\t$ are $f$-consistent, and that $T(f) := g$ is a transport process (TP) with law denoted as $T\#\eta := \pi$.
\end{definition}

The main idea of the previous definition is to start from a simple stochastic process, one that is easy to simulate, and then to generate another stochastic process that is more complex and more expressive. Since our purpose is to model data through their finite-dimensional laws, our definition implies a correspondence between the laws of the reference process and those of the objective process; for this reason, it is important that the mappings retain the size of the distributions and the respective indexes.

It is straightforward that are many collection of measurable maps that are inconsistent, even in some simple cases. For example, consider the swap maps given by $T_1(x_1) = x_1$, $T_{12}(x_1,x_2) = (x_2,x_1)$ and so on. If $f$ is a heteroscedastic Gaussian process, then we have $F_1(x_1) = \N_1(x_1|0,\sigma_1^2)$ and $F_{12}(x_1, x_2)=\N_2\left((x_1, x_2)|0,\begin{bmatrix} \sigma_1^2 & \sigma_{12} \\ \sigma_{12} & \sigma_2^2 \end{bmatrix}\right)$. The push-forward distributions are given by $G_1(y_1)=\N_1(x_1|0,\sigma_1^2)$ and $G_{12}(y_1,y_2)=\N_2\left((y_1, y_2)|0,\begin{bmatrix} \sigma_2^2 & \sigma_{12} \\ \sigma_{12} & \sigma_1^2 \end{bmatrix}\right)$, and since $\lim\limits_{y_2 \to \infty} G_{12}(y_1,y_2) = \N_1(x_1|0,\sigma_2^2) \neq \N_1(x_1|0,\sigma_1^2) = G_1(y_1)$, so we have that $T$ is inconsistent for $f$. Note that if $f$ is a trivial \emph{i.i.d.} stochastic process, then $T$ is $f$-consistent.

To be able to use transport processes as regression models, we must be able to define a finitely-parameterised transport $T^\theta$ with $\theta \in \Theta \subset \R^d$, where the finite-dimensional maps $(T^\theta)_\t$ are consistent and invertible. For example, given $\theta \in \Theta = \X$ the \emph{shift} transport is $T^\theta =\{T_{\t}(\x) = \x + \theta | \t \in \T^n, n \in \mathbb{N}\}$, or simply $(T^\theta)_{\t}(\x) = \x + \theta $.  For simplicity, if there is no ambiguity, we will denote $(T^\theta)_\t$ as $T_\t$. In the next sections, we will show more sophisticated examples of finitely-parameterised transports $T^\theta$, so in what follows we concentrate on explaining the general approach of using TP as regression models.

\subsection{Learning transport process}

As in the GP approach, given observations, the learning task corresponds to finding the \emph{best} transport $T^\theta$, determined by the parameters $\theta$ that minimises the negative logarithm of their marginal likelihood (NLL), given below.
\begin{proposition}
	Let $g = T^\theta(f)$ be a transport process with law $\pi = T^\theta\#\eta$, where $\eta$ has finite-dimensional distributions with density denoted $\eta_{\t}$. Given observations $(\t,\y)$, if the map $T_\t$ is invertible on $\y$ (for simplicity we denote $T_\t^{-1}$ as $S_\t$) and differentiable on $\x=S_\t(\y)$, its NLL is given by
	\begin{align}
	\label{eq:TGP_NLL}
	-\log \pi_{\t}(\y|\theta) &= -\log \eta_{\t}(S_{\t}(\y)) - \log |\nabla S_{\t}(\y)|\nonumber\\
	&= -\log \eta_{\t}(S_{\t}(\y)) + \log |\nabla T_{\t}(S_{\t}(\y))|.
	\end{align}
\end{proposition}
The first equality is due to the change of variables formula \cite{hogg1995introduction}. For the second identity, via the inverse function theorem \cite{rudin1964principles} we have that $\nabla S_{\t}(\y) = \nabla T_{\t}(\x)^{-1}$, and by the determinant of the inverse property \cite{petersen2008matrix} we get $|\nabla T_{\t}(\x)^{-1}| = |\nabla T_{\t}(\x)|^{-1}$. To calculate eq.~\eqref{eq:TGP_NLL} we need to be able to compute the log-density of $\eta_{\t}$, the inverse $S_{\t}$, and the gradient $\nabla T_{\t}$ (or $\nabla S_{\t}$).

It is important to note that the reference process is fixed and the trainable object corresponds to transport. In other words, following the principle known as \emph{reparametrisation trick} \cite{kingma2013auto}, the model is defined so that random sources have no parameters, so that optimization algorithms can be applied over deterministic parametric functions. Akin to the GP approach, the NLL for transport process (eq.~\eqref{eq:TGP_NLL}) follows an elegant interpretation of how to avoid overfitting:
\begin{itemize}
	\item The first term $-\log \eta_{\t}(S_{\t}(\y))$ is the \emph{goodness of fit} score between the model and the data, privileging those $\theta$ that make $S_{\t}(\y)$ to be close to the mode of $\eta_{\t}$. E.g., if $\eta_{\t}$ is a standard Gaussian, this term (omitting a constant) is $\frac{1}{2}\lVert S_{\t}(\y) \rVert_{2}^2$, and with enough observations it results in overfitting: $S_{\t}$ is the null function.
	\item On the other hand, the second term $-\log |\nabla(S_{\t}(\y)|$ is the \emph{model complexity penalty}, and it prioritises those $\theta$ that make $|\nabla S_{\t}(\y)|$ to be large, i.e. $S_{\t}$ has large deviations around $\y$, thus avoiding the null function and, in turn, the overfitting. Note that a valid map satisfies $|\nabla S_{\t}(\y)|>0$.
\end{itemize}  

\subsection{Inference with transport process}

Once the transport $T^\theta$ is trained, via minimising the NLL, inference is performed via calculating the posterior distribution of $(\tto,\yo)$ given observations $(\t,\y)$ under the law $\pi$: for any inputs $\tto$ we compute the posterior distributions $\pi_{\tto|\t}(\cdot|\y)$. As our goal is to generate stochastic processes more expressive than GPs, the mean and variance are not sufficient to compute (e.g. we need expectations associated with extreme values). For this reason, our approach is based on generating efficiently independent samples from $\pi_{\tto|\t}$, to then perform calculations via Monte Carlo methods \cite{rubinstein2016simulation}. 

Since we assume that we can easily obtain samples from $\eta_{\tto}$ (and  $\eta_{\tto|\t}$ if necessary), we will show how to use these samples and the transport $T^\theta$ to efficiently generate samples from $\pi_{\tto|\t}$. The principle behind this idea is that if $\pi_{\tto|\t} = \varphi\#\eta_{\tto}$ and $\x \sim \eta_{\tto}$ then $\varphi(\x) \sim \pi_{\tto|\t}$. In cases where this principle can not be applied, we can alternatively obtain samples using methods based on MCMC, which need to be able to evaluate the density of the posterior distribution.

\section{Marginal Transport}
\label{sec:marginaltransport}

In this section, we present a family of transports named \emph{marginal transports}, given that they can change the marginals distributions of a stochastic process, extending in this way the mean function from GPs, as well as the warping function from WGPs, including the model CWGP presented previously on Chapter \ref{sec:cwgp}. We prove their consistency, deliver the formulas for training, and give a general method to sampling.

\begin{definition}
	$T=\{T_\t | \t \in \T^n, n \in \mathbb{N}\}$ is a marginal transport if there exists a measurable function $h: \T \times \X \rightarrow \X$, so that $[T_\t(\x)]_i = h(t_i, x_i)$ for $\t \in \T^n, \x \in \X^n, n \in \mathbb{N}$. Additionally, if $h(t,\cdot):\X \rightarrow \X$ is increasing (so differentiable a.e.) for all $t \in \T$, then we said that $T$ is a increasing marginal transport.
\end{definition}

A \emph{marginal} transport is defined in a \emph{coordinate-wise} manner via the function $h$. For example, given a \emph{location} function $m: \I \rightarrow \X$, then $h(t,x) = m(t)+x$ induces a marginal transport $T^h$ such that if $\eta = \GP(0,k)$ then $T^h\#\eta = \GP(m, k)$. As $T^h$ determinates the mean on the induced stochastic process, usual choices for $m$ are elementary functions like polynomial, exponential, trigonometric and additive/multiplicative combinations.

However, this family of transports is more expressive than just determining the mean, being able to define higher moments such as variance, skewness and kurtosis. This expressiveness can be achieved, beside the \emph{location} function $m$, by considering a \emph{warping} $\varphi: \Y \rightarrow \X$ to define the transport $T^h$ induced by the composite function $h(t,x) = \varphi^{-1}\left(m(t)+x\right)$, such that if $\eta = \GP(0,k)$ then we have that $T^{h}\#\eta = \WGP(\varphi, m, k)$. The most common \emph{warping} functions are affine, logarithm, Box-Cox \cite{rios2018learning}, and sinh-arcsinh \cite{Sinharcsinh}, which can be composed to generate more expressive warpings. This layers-based model, named compositionally WGP, has been thoroughly studied in previous works \cite{rios2018learning, riostobar2019cwgp}. However, the expressiveness of marginal transport is more general since the warping function can change across the coordinates.

\subsection{Consistency of the marginal transport}

Marginal transports are well-defined with a GP reference, in the sense that it always defines a set of consistent finite-dimensional distributions, and thus it induces a stochastic process. The following proposition shows that this family of transports is compatible with any stochastic process, a property which we refer to as \emph{universally consistent}.

\begin{proposition}
	Given any stochastic process $f=\left\{ x_{t}\right\} _{t\in \T}$ and any increasing marginal transport $T$, then $T$ is an $f$-transport.
	\begin{proof}
		Given $\eta_{\t} \in \mathcal{F}$ a  finite-dimensional distribution, the transported cumulative distribution function is given by $F_{\pi_{\t}}(\y) = F_{\eta_{\t}}((h^{-1}(t_i,y_i))_{i=1}^n)$, where $h^{-1}(t,\cdot)$ denotes the inverse on the $\X$-coordinate of $h$, which is also increasing. 
		
		The marginalisation condition is fulfilled since $F_{\eta_{\t,t_{n+1}}}(\x,\infty) = F_{\eta_{\t}}(\x)$, so we have
		\begin{align*}
		F_{\pi_{\t,t_{n+1}}}(\y,\infty) &= F_{\eta_{\t,t_{n+1}}}((h^{-1}(t_i,y_i))_{i=1}^n, h^{-1}(t_{n+1},\infty)),\\
		&=F_{\eta_{\t,t_{n+1}}}((h^{-1}(t_i,y_i))_{i=1}^n, \infty)
		=F_{\eta_{\t}}((h^{-1}(t_i,y_i))_{i=1}^n)=F_{\pi_{\t}}(\y).
		\end{align*}
		
		Given an $n$-permutation $\tau$, we denote $\tau(\t) = t_{\tau(1)},...,t_{\tau(n)}$ and $\tau(\y) = y_{\tau(1)},...,y_{\tau(n)}$. Since $F_{\eta_{\tau(\t)}}(\tau(\x)) = F_{\eta_{\t}}(\x)$ then  $F_{\pi_{\tau(\t)}}(\tau(\y)) =  F_{\eta_{\tau(\t)}}((h^{-1}(t_{\tau(i)},y_{\tau(i)}))_{i=1}^n)=F_{\eta_{\t}}((h^{-1}(t_i,y_i))_{i=1}^n)=F_{\pi_{\t}}(\y)$, satisfying the conditions. 
	\end{proof}
\end{proposition}

\begin{remark}
	In general we will assume that marginal transports are increasing, due to for any fixed stochastic process $f$ and any marginal transport $T$, exist an increasing marginal transport $T^h$ such that $T\#f$ and $T^h\#f$ have the same distributions (i.e. all their finite-dimensional distributions agree \cite{shalizi2010almost}). The increasing function $h$ is defined via the unique monotone transport maps from $\eta_t$ to $\pi_t$ given by $h(t,x) = F_{\pi_t}^{-1}(F_{\eta_t}(x))$ for each $t \in \T$ \cite{cuestaalbertos1993optimal}.
\end{remark}

Marginal transports $T^h$ satisfy straightforwardly the consistency condition since there are coordinate-wise maps. This \emph{diagonality} is an appealing mathematical property, but it has a high cost: the transport process inherits the same copula from the reference process. This fact implies that independent marginals, such as white noise, remain independent with the marginal transport. The following proposition shows the benefits and limitations of diagonality \cite{wilson2010copula}.

\begin{proposition}
	Let $f= \{x_t\}_{t\in\T}$ be a stochastic process with marginal cumulative distribution functions $F_t$ for $t \in \T$, and copula process $C$. Given any sequence of cumulative distribution functions $\{G_t\}_{t \in \I}$, the function $h(t,x) = G_t^{-1}(F_t(x))$ induces a marginal transport $T^h$ where $g = T^h\#f$ is a transport process with marginals $G_t$ and copula process $C$.
	\begin{proof}
		The copula of $f$ is the stochastic process $C = \{C_t\}_{t\in\T}$ where $C_t := F_t(x_t)$ follows a uniform distribution. The transport process $g = T^h\#f = \{y_t\}_{t\in\T}$ satisfies $y_t = G_t^{-1}(F_t(x_t)) = G_t^{-1}(C_t)$, so its copula process $D = \{D_t\}_{t\in\T}$ is given by $D_t = G_t(y_t) = G_t(G_t^{-1}(C_t)) = C_t$. Thus, $f$ and $g$ have the same copula.
	\end{proof}
\end{proposition}

\subsection{Learning of the marginal transport}

For learning we have to calculate the NLL given by eq. \eqref{eq:TGP_NLL}. The inverse map is given by $S_{\t}(\y)_{i} = h^{-1}(t_i, y_i) = x_i$ and the \emph{model complexity penalty} is given by
\begin{align}
\log|\nabla S_{\t}(\y)| = \sum_{i} \log \frac{\partial h^{-1}}{\partial  y}(t_i, y_i)=- \sum_{i} \log \frac{\partial h}{\partial  y}(t_i, x_i).
\end{align}
E.g., if $h(t,x) = \varphi^{-1}\left(m(t)+\sigma(t)x\right)$, then $h(t,y)^{-1} = \frac{\varphi(y) -  m(t)}{\sigma(t)}$ and $\log|\nabla S_{\t}(\y)| = \sum_{i} \log \frac{\varphi'(y_i)}{\sigma(t_i)}$.

\subsection{Inference with marginal transport}

For inference on new inputs $\tto$, the posterior distribution $\pi_{\tto|\t}(\cdot|\y)$ is the push-forward of $\eta_{\tto|\t}(\cdot|S_{\t}(\y))$ by $T_{\tto}$, so if $\xo \sim \eta_{\tto|\t}(\cdot|S_{\t}(\y))$ then $\yo=T_{\tto}(\xo) \sim \pi_{\tto|\t}(\yo|\y)$. Note that the probability of a set $E$ under the density of $\pi_t$ is equal to the probability of the image $h_t^{-1}(E)$ under the density of $\eta_t$, where $h_t(\cdot) :=h_\theta(t, \cdot)$. Thus, if we can compute marginals quantiles under $\eta_\t$, such as the median and confidence intervals, we can do the same under $\pi_\t$. Even more, the expectation of any measurable function $v:\mathcal{Y}\rightarrow \mathbb{R}$ under the law $\pi_{\t}(\y)$ is given by $\mathbb{E}_{\pi_{\t}}\left[ v\left(\y\right)\right]  =\mathbb{E}_{\eta_{\t}}\left[v\left(h_{\t}\left( \x\right)\right) \right]$.

\section{Covariance Transport}
\label{sec:covariancetransport}

From the results of the previous section, the only way to induce a different copula under our transport-based approach is to consider non-diagonal maps. The problem with these maps is that we lose the property of \emph{universally consistent}, but it is possible to find conditions over the reference stochastic processes so that the transport is consistent. 

In this section, we present a family of transports named \emph{covariance transports}, that allows us to change the covariance, and therefore the correlation, over the induced stochastic process. These transports are based on covariance kernels, e.g. the squared exponential given by $k(t,s) = \sigma^2\exp(-r|t-s|^2)$ with parameters $\theta=(\sigma,r)$. 

\begin{definition}
	$T^k=\{T_\t | \t \in \T^n, n \in \mathbb{N}\}$ is a covariance transport if there exists a covariance kernel $k:\T \times \T \rightarrow \R$, so that $T_{\t}(\x) = L_{\t}\x$, where $L_{\t}$ is a square root of $\Sigma_{\t\t} = k(\t,\t)$, i.e. $L_{\t}L_{\t}^\top = \Sigma_{\t\t}$.
\end{definition}

Since $\Sigma_{\t\t}$ is a definite positive matrix, always exist an unique definite positive square root denoted $\Sigma_{\t\t}^{1/2}$ and named the \emph{principal square root} of $\Sigma_{\t\t}$. Additionally, always exist an unique lower triangular square root denoted $\chol(\Sigma)$ and named as the \emph{lower Cholesky decomposition} of $\Sigma_{\t\t}$, where later we will show his importance to getting practical transports.

If $T^k$ is a covariance transport induced by $k$ and $f \sim \GP(0, \delta(t,\bar t))$ is a Gaussian white noise process, then we have that $T^k$ is a $f$-transport where $T^k(f) \sim \GP(0, k)$, i.e. $T^k$ fully defines the covariance over the transport process. This fact is true due to the maps $T_\t(\x)$ being linear (given by $T_\t(\x)_i = \sum_{j=1}^n l_{ij}x_j$ where $[L_\t]_{ij}=l_{ij}$), so given a finite-dimensional law $\eta_{\t} = \sim \N_n(0,I)$, by the linear closure of Gaussian distributions we have that $T_{\t}\#\eta_{\t} = \N_n(0,\Sigma_{\t\t})$ where $L_{\t}L_{\t}^{\top} = \Sigma_{\t\t} = k_{\theta}(\t,\t)$. We assume for now the consistency of the covariance transport, but we will study it at the end of this section, once we have revised the concept of triangularity.

\subsection{Learning of the covariance transport}

We say that a finite-dimensional map $T_{\t}: \R^n \to \R^n$ is \emph{triangular} if it structure is triangular, in the sense $T_\t(\x)_i = T_i(x_1,...,x_i)$ for $i=1,...,n$. If $T_{\t}$ is differentiable, then it is triangular if and only if its Jacobian $\nabla T_{\t}$ is a lower triangular matrix. We say that a transport $T$ is \emph{triangular} if its finite-dimensional maps are triangular. While a marginal transport is \emph{diagonal}, a covariance transport with lower Cholesky decomposition is \emph{triangular}. Note that diagonal maps are also triangular maps, and the composition of triangular maps remains triangular. Triangularity is an appealing property for maps, since it allows us to perform calculations more efficiently that in the general case. The following result shows the similarity between triangular and diagonal maps for the learning task.

\begin{proposition}
	Let $T_\t$ be an invertible and differentiable triangular map on $\x$. If we denote $T_\t(\x) = \y$ then:
	\begin{itemize}
		\item the inverse map $S_\t$ is also triangular that fulfills that $S_\t(\y) = \x$,
		\item the model complexity penalty is given by $$\log|\nabla S_{\t}(\y)| = \sum_{i} \log \frac{\partial S_i}{\partial  y_i}(y_1,...,y_i)=- \sum_{i} \log \frac{\partial T_i}{\partial  x_i}(x_1,...,x_i).$$
	\end{itemize}
	\begin{proof}
		The first coordinate satisfies $T_1(x_1) = y_1$ so $S_1(y_1) = x_1$. By induction, we have $S_k(y_1,...,y_k) = x_k$, and since $T_{k+1}(x_1,...,x_{k+1}) = y_{k+1}$, then we have the equation $$T_{k+1}(S_1(y_1),...,S_k(y_1,...,y_k), x_{k+1}) = y_{k+1},$$ so we can express $x_{k+1}$ in function of $y_1,...,y_{k+1}$, i.e. $S_{k+1}(y_1,...,y_{k+1}) = x_{k+1}$ so $S_{\t}$ is triangular. With this we have that $\nabla S_{\t}(\y)$ is a lower triangular matrix, so its determinant is equal to the product of all the elements on the diagonal. The complexity penalty, then, is analogous to the diagonal case.
	\end{proof} 	
\end{proposition}
For triangular covariance transports we have that $S_\t(\y)=L_{\t}^{-1}\y$, which can be computed straightforwardly via forward substitution \cite{demmel1997applied}, and $\log|\nabla S_{\t}(\y)| = - \sum_{i} \log l_{ii}$, where $l_{ii}$ are the diagonal values of $L_{\t}$. 

\subsection{Inference with the covariance transport}
Triangular maps allow efficient inference since posterior distributions can be calculated as a push-forward from the reference.
\begin{proposition}
	\label{prop_triangular}
	Given observations $\y \sim \pi_{\t}$, denote $\x = T_{\t}^{-1}(\y)$ and by $\eta_{\tto|\t}(\xo|\x)$ the posterior distribution of $\eta$. Assume that the transports $T_{\t}$ are triangular, then the posterior distribution of $\pi$ is given by
	\begin{align}
	\pi_{\tto|\t}(\yo|\y) = \left[P_{\tto} \circ T_{\t,\tto}^{\x}\right]\#\eta_{\tto|\t}(\cdot|\x),
	\end{align}
	where $T_{\t,\tto}^{\x}(\cdot) = T_{\t,\tto}(\x,\cdot)$, and $P_{\tto}(\cdot)$ is the projection on $\tto$, i.e. $P_{\tto}(\x, \xo) = \xo$.
	\begin{proof}
		Since the maps are triangular, their inverses also are triangular: $$T_{\t,\tto}^{-1}(\y, \yo) = [T_{\t}^{-1}(\y), T_{\tto|\t}^{-1}(\yo|T_{\t}^{-1}(\y))],$$ and as its gradient it is also triangular, then their determinants satisfy $$|\nabla T_{\t,\tto}^{-1}(\y, \yo)| = |\nabla T_{\t}^{-1}(\y)| |\nabla_{\yo} T_{\tto|\t}^{-1}(\yo|T_{\t}^{-1}(\y))|.$$ With these identities, the posterior density of $\pi_{\tto|\t}(\yo|\y)$ is given by
		\begin{align*}
		\pi_{\tto|\t}(\yo|\y) =& \frac{\pi_{\t, \tto}(\y, \yo)}{\pi_{\t}(\y)} =  \frac{\eta_{\t, \tto}(T_{\t,\tto}^{-1}(\y, \yo))|\nabla T_{\t, \tto}^{-1}( \y, \yo)|}{\eta_{\t}(T_{\t}^{-1}(\y))|\nabla T_{\t}^{-1}(\y)|},\\
		=& \frac{\eta_{\t,\tto}(T_{\t}^{-1}(\y),T_{\tto|\t}^{-1}(\yo|T_{\t}^{-1}(\y)))}{\eta_{\t}(T_{\t}^{-1}(\y))} \frac{|\nabla T_{\t}^{-1}(\y)| |\nabla_{\yo} T_{\tto|\t}^{-1}(\yo|T_{\t}^{-1}(\y))|}{|\nabla T_{\t}^{-1}(\y)|},\\ =& \eta_{\tto|\t}(T_{\tto|\t}^{-1}(\yo|T_{\t}^{-1}(\y))|T_{\t}^{-1}(\y)) |\nabla_{\yo} T_{\tto|\t}^{-1}(\yo|T_{\t}^{-1}(\y))|, \\
		=& T_{\t,\tto}(T_{\t}^{-1}(\y), \cdot)|_{\tto}\#\eta_{\tto|\t}(\cdot|T_{\t}^{-1}(\y))
		= \left[P_{\tto} \circ T_{\t,\tto}^{\x}\right]\#\eta_{\tto|\t}(\cdot|\x).
		\end{align*}
	\end{proof}	
	
\end{proposition}

For the covariance transport, and given new inputs $\tto$, the posterior distribution $\pi_{\tto|\t}(\yo|\y)$ is the push-forward of $\eta_{\tto|\t}(\cdot|L_{\t}^{-1}\y)$ by the affine map $T(\u) = A_{\t}L_{\t}^{-1}\y + A_{\tto}\u$, where $L_{\t,\tto}=\left[ \begin{array}{cc} L_{\t} & 0\\ A_{\t} & A_{\tto} \end{array}\right]$. Note that $A_{\t}L_{\t}^{-1} = \Sigma_{\tto\t}\Sigma_{\t\t}^{-1}$ and $A_{\tto}A_{\tto}^\top = \Sigma_{\tto\tto} - \Sigma_{\tto\t}\Sigma_{\t\t}^{-1}\Sigma_{\tto\t}$, so the map agrees with $T(\u) = \Sigma_{\tto\t}\Sigma_{\t\t}^{-1}\y + L_{\tto|\t}\u$, where $L_{\tto|\t} = \chol(\Sigma_{\tto|\t})$ with $\Sigma_{\tto|\t} = \Sigma_{\tto\tto} - \Sigma_{\tto\t}\Sigma_{\t\t}^{-1}\Sigma_{\tto\t}$. 

\subsection{Consistency of the covariance transport}

Going back to the issue of consistency, the following proposition gives us a condition over triangular maps that imply consistency under marginalisation.
\begin{proposition}
	Let $T=\{T_\t:\X^n \to \X^n | \t \in \T^n, n \in \mathbb{N}\}$ be a collection of triangular measurable maps that satisfy $P_{\t} \circ T_{\t,t_{n+1}}(\y,y_{n+1}) = T_{\t}(\y)$, with $P_{\t}$ the projection on $\t$. Then $T$ is universally consistent under marginalisation.
	\begin{proof}
		The push-forward finite-dimensional distribution function is $F_{\pi_{\t}}(\y) = F_{\eta_{\t}}(S_{\t}(\y))$. Since a valid map satisfies $\frac{\partial S_i}{\partial  y_i}(y_1,...,y_i)>0$ for all $i\ge 1$, then $S_{t_{n+1}}$ is increasing on $y_{n+1}$ so $S_{t_{n+1}}(\y,\infty) = \infty$. With this, if $P_{\t} \circ T_{\t,t_{n+1}}(\y,y_{n+1}) = T_{\t}(\y)$ then the inverse also satisfies this. Finally, the marginalisation condition is fulfilled becauses $F_{\pi_{\t,t_{n+1}}}(\y,\infty) = F_{\eta_{\t,t_{n+1}}}(S_{\t,t_{n+1}}(\y,\infty)) = F_{\eta_{\t,t_{n+1}}}(S_{\t}(\y), S_{t_{n+1}}(\y,\infty)) = F_{\eta_{\t,t_{n+1}}}(S_{\t}(\y), \infty) = F_{\eta_{\t}}(S_{\t}(\y)) = F_{\pi_{\t}}(\y)$.
	\end{proof}
\end{proposition}

Note that diagonal and covariance transports satisfy the above condition, that can be interpreted like an \emph{order} between their finite-dimensional triangular maps. The consistency under permutations means that, given any $n$-permutation $\tau$, it satisfies $F_{\pi_{\tau(\t)}}(\tau(\y)) = F_{\pi_{\t}}(\y)$, or equivalently, $F_{\eta_{\tau(\t)}}(S_{\tau(\t)}(\tau(\y))) = F_{\eta_{\t}}(S_{\t}(\y))$. Since $\eta$ is consistent under permutations, we have the following condition over $\eta_{\t}$ and $S_{\t}$:
\begin{align}
F_{\eta_{\t}}(\tau^{-1}(S_{\tau(\t)}(\tau(\y)))) = F_{\eta_{\t}}(S_{\t}(\y)).
\end{align}
The above equality can be written in terms of the density function as
\begin{align}
\eta_{\t}(\tau^{-1}(S_{\tau(\t)}(\tau(\y)))) \lv \nabla( \tau^{-1}(S_{\tau(\t)}(\tau(\y)))) \rv = \eta_{\t}(S_{\t}(\y)) \lv \nabla S_{\t}(\y) \rv.
\end{align}
Note that if $T$ is \emph{universally} consistent under permutations, then it has to satisfy $\tau(S_{\t}(\y)) = S_{\tau(\t)}(\tau(\y)))$, so $T$ must be diagonal. This mean that strictly triangular transports can be consistent only for some families of distributions. The following proposition shows one condition over $\eta$ for consistency of covariance transports.

\begin{proposition}
	Let $f=\left\{ x_{t}\right\} _{t\in \T}$ be a stochastic process where its finite-dimensional laws have densities with the form $\eta_{\t}(\x) = \beta_n(\xn)$, for some functions $\beta_n$ with $n = |\t|$. Then any triangular covariance transport $T^k$ is an $f$-transport.
	\begin{proof}
		We just need to check consistency under permutations. We have that $S_\t(\y)=L_{\t}^{-1}\y$, so $|\nabla S_{\t}(\y)| = |L_{\t}|^{-1} = \prod_{i} l_{ii}^{-1}$, where $l_{ii}$ are the diagonal values of $L_{\t}$. Note that this calculation is independent of $\y$ and it only depends on the values of the diagonal, so $\lv \nabla( \tau^{-1}(S_{\tau(\t)}(\tau(\y)))) \rv = |L_{\tau(\t)}|^{-1} =  \prod_{i} d_{ii}^{-1}$, where $d_{ii}$ are the diagonal values of $L_{\tau(\t)}$. Since $|\Sigma_{\t\t}| = |L_{\t}|^2$ and $|\Sigma_{\tau(\t)\tau(\t)}| = |P_{\tau}\Sigma_{\t\t}P_{\tau}| = |\Sigma_{\t\t}|$ then we have that $|L_{\tau(\t)}| = |L_{\t}|$. With this identity, we need that $\eta_{\t}(\tau^{-1}(L_{\tau(\t)}^{-1}\tau(\y)))  = \eta_{\t}(L_{\t}^{-1}\y)$, but this is fulfilled under the hypothesis over $\eta_{\t}$, since
		\begin{align*}
		\eta_{\t}(\tau^{-1}(S_{\tau(\t)}(\tau(\y)))) = \beta_n\left( \lV \tau^{-1}(L_{\tau(\t)}^{-1}\tau(\y))\rV_2 \right) = \beta_n\left( \tau(\y)^{\top}\Sigma_{\tau(\t)\tau(\t)}^{-1}\tau(\y)) \right)\\ = \beta_n\left(\y\Sigma_{\t\t}^{-1}\y\right) = \eta_{\t}(L_{\t}^{-1}\y).
		\end{align*}
	\end{proof}
\end{proposition}

Note that the standard Gaussian distribution satisfies the hypothesis with $\beta_n(r)= c_n\exp(-r^2/2)$ where $c_n =(2\pi)^{-n/2}$. This family of distributions is known in the literature as spherical distributions, and their generalisation with covariance is known as elliptical distributions \cite{owen1983class}. In the next section, we will study these distributions via a new type of transports.

\section{Radial Transports}
\label{sec:familytgp}

While covariance and marginal transports can model correlation and marginals, they inherit the base copula from the reference. For example, if the reference process is a GP, through covariance and marginal transports we can only generate WGP with Gaussian copulas. Our proposal to construct other copulas relies on radial transformations that are capable of modifying the norm of a random vector, changing its copula in this way.

\begin{definition}
	$T=\{T_\t | \t \in \T^n, n \in \mathbb{N}\}$ is a radial transport if there exists a radial function $\phi(r) = \frac{\alpha(r)}{r}$, with $\alpha:\R^{+}\to\R^{+}$ monotonically non-decreasing, and $\lV \cdot \rV$ a norm over $\X^n$ so that $T_{\t}(\x) = \phi(\lVert \x \rVert)\x$.
\end{definition}

According to the chosen norm $\lV \cdot \rV$, the copula family generated by our approach is different. The Euclidean $\ell_2$ norm, $\lV \cdot \rV_2$, allows us to define elliptical processes; the Manhattan $\ell_1$ norm, $\lV \cdot \rV_1$, allows us to define Archimedean processes. In the following sections we will study these respective \emph{elliptical transports} and \emph{Archimedean transports}.

\subsection{Elliptical processes}

In the previous section, we introduced a particular family of distributions known as spherical distributions that are consistent with covariance transport. We now introduce a generalisation called elliptical distributions \cite{owen1983class}.

\begin{definition}
	$\x \in \R^n$ is elliptically distributed iff there exists a vector $\mu \in \R^n$, a (symmetric) full rank scale matrix $A \in \R^{n \times n}$, a uniform random variable $U^{(n)}$ on the unit sphere in $\R^{n}$, i.e. $\lV U^{(n)} \rV_2 =1$, and a real non-negative random variable $R \in \R^+$, independent of $U^{(n)}$, such that $\x \stackrel{d}{=} \mu + RAU^{(n)}$, where $\stackrel{d}{=}$ denotes equality in distribution. 
\end{definition}
\begin{remark}
	If $\x$ is elliptically distributed and has density $\eta(\x)$, then for some positive function $\beta_n$, it has the form $\eta(\x) = \lv \Sigma\rv^{-1/2} \beta_n((\x-\mu)^\top\Sigma^{-1}(\x-\mu))$, where  $\Sigma = A^{\top}A$ and $R$ has density $p_R(r) = \frac{2\pi^{n/2}}{\Gamma(n/2)}r^{n-1}\beta_n(r^2)$ \cite{owen1983class}.	
\end{remark}
Gaussian distributions are members of elliptical distributions: if $\x \sim \N_{n}(0,\Sigma_{\x\x})$ then $\x \stackrel{d}{=}  R_{n}L_{\t}U^{(n)}$ with $R_{n} \sim \sqrt{\chi^{2}(n)}$ (i.e. follow a Rayleigh distribution) and $\Sigma_{\x\x}=L_{\t}^\top L_{\t}$. However elliptical distributions include other distributions like the Student-t \cite{demarta2005t}, a widely-used alternative due to its heavy-tail behaviour. Elliptical processes have a useful characterisation as follows:
\begin{theorem}[Kelker's theorem \cite{kelker1970distribution}]
	$f$ is an elliptical process where the finite-dimensional marginals $\x$ have density if and only if there exists a positive random variable $R$ such that $\x|R \sim \N_{n}(\mu_{\x},R\Sigma_{\x\x})$.
\end{theorem}
The above result can be summarised in that elliptic processes are mixtures of Gaussian processes. This characterisation gives us a direction to achieve our goal through radial transports.

\subsubsection{Elliptical transport}

Our goal is to define stochastic processes via our transport approach where their copula is elliptical, beyond the Gaussian case. Let us set some notation. Given a r.v. $R$, its cumulative distribution function is denoted $F_R$. The square-root of a chi-squared (a.k.a. Rayleigh) distributed r.v. will be denoted $R_{n} \sim \sqrt{\chi^{2}(n)}$. Our idea to transport a Gaussian copula to another elliptical copula is based on the following optimal transport result \cite{cuestaalbertos1993optimal, ghaffari2018multivariate}.

\begin{proposition}
	Let $\x \stackrel{d}{=} RAU^{(n)}$ be an elliptically distributed r.v. Given a positive r.v. $S$, consider the radial map $T^\alpha(\x) = \phi(\xn)\x = \frac{\alpha(\lVert A^{-1}\x \rVert_2)}{\lVert A^{-1}\x \rVert_2}\x$ where $\alpha(r) = F_{S}^{-1}(F_R(r))$. Then we have that $T^\alpha(\x) \stackrel{d}{=} SAU^{(n)}$.
\end{proposition}

A useful property of this type of transports is that we can generate distributions with different elliptical copulas by changing the norm without altering the correlation.
\begin{lemma}
	The radial transport $T^\alpha$ does not modify the correlation.
	\begin{proof}
		Let $\x \stackrel{d}{=} RAU^{(n)}$. Then, $Cov(\x) = \frac{\E(R^2)}{rank(A)}A^\top A = c\Sigma$. As $\y =: T_{\t}(\x) \stackrel{d}{=}  \alpha(R)AU^{(n)}$ then $Cov(\y) = \frac{\E(\alpha(R)^2)}{rank(A)}A^\top A = d\Sigma$. As $Cov(\y) = \frac{d}{c}Cov(\x)$, we have $Corr(\y) = Corr(\x)$. 
	\end{proof}
\end{lemma}

Note that if $\x \stackrel{d}{=} RU^{(n)}$ then $T^\alpha(\x) = \phi(\xn)A\x \stackrel{d}{=} \alpha(R)AU^{(n)}$ . Since we can decompose $T^\alpha(\x) = A(\phi(\xn)\x)$ in a covariance transport, we merely consider the elliptical transport as $T_{\t}(\x) = \phi(\xn)\x$. The next result characterises a family of transports based on radial functions that generate elliptical processes from Gaussian white noise processes.

\begin{theorem}
	Let $p_\theta$ be a density function supported on positive real line. Define $F_{R_{n,\theta}}(r) := \int_{0}^{\infty}p_{\theta}(s)F_{R_n}(r/s)ds$ and $\alpha_{n,\theta}(r) = F^{-1}_{R_{n,\theta}} \circ F_{R_n}(r)$. Then the elliptical radial transport defined by $T_{\t}(\x): = \frac{\alpha_{n,\theta}(\lVert \x\rVert_{2})}{\lVert \x\rVert_{2}}\x$ 
	is an $f$-transport with $f \sim \GP(0,\delta(t,\bar t))$, where the transport process $g := T(f)$ has finite-dimensional elliptical distributions.
	\begin{proof}
		Let $R_{\theta}$ be a positive r.v. with density function  $p_\theta$. Since $R_{n} \sim \sqrt{\chi^{2}(n)}$ is also a positive r.v., by the product distribution formula \cite{rohatgiintroduction} we have that the r.v. $R_{n,\theta} := R_{\theta} R_{n}$ has a cumulative distribution function given by $F_{R_{n,\theta}}(r) := \int_{0}^{\infty}p_{\theta}(s)F_{R_n}(r/s)ds$. Given that the finite-dimensional laws of $f$ are $\eta_{\t} = \N_n(0,I)$, if $\x \sim \eta_{\t}$, then $\xn \stackrel{d}{=} R_n$, so $\alpha_{n,\theta}(\xn) \stackrel{d}{=} R_{n,\theta} \stackrel{d}{=} R_{\theta} R_{n}$ and $\frac{\x}{\xn}  \stackrel{d}{=} U^{(n)}$ are independent, having thus that $T_{\t}(\x) \stackrel{d}{=} R_{\theta} R_{n} U^{(n)}$ is elliptically distributed. Since $T_{\t}(\x)|R_{\theta} \sim \N_{n}(0,R_{\theta}^2I)$ and $R_{\theta}$ is independent of $\x$, by Kelker's theorem the push-forward finite-dimensional distributions $ \mathcal{\hat F} = \{T_{\t}\#\eta_{\t} | \t \in \T^n, n \in \mathbb{N}\}$ are consistent and define an elliptical process.
	\end{proof}
\end{theorem}

\subsubsection{Learning of the elliptical transport}

The following proposition allow us to calculate the determinant of the gradient of this radial transport.
\begin{proposition}
	Let $T_{\t}(\x)= \phi(\xn)\x = \frac{\alpha(\xn)}{\xn}\x$. Then $|\nabla T_{\t}(\x)| = \phi(\left\Vert \x \right\Vert_{2})^{n-1}\alpha'(\xn)$. 
	\begin{proof}
		\begin{align*}
		\frac{\partial T_{\t}(\x)_i}{\partial x_i} &= \phi(\xn) + \phi'(\xn)\frac{x_i^2}{\lVert \x \rVert_2},\\
		\frac{\partial T_{\t}(\x)_i}{\partial x_j} &= \phi'(\xn)\frac{x_i x_j}{\lVert \x \rVert_2}, \text{if } i\neq j,\\
		\nabla T_{\t}(\x) &= \frac{\phi'(\left\Vert \x \right\Vert_{2})}{\left\Vert \x \right\Vert_{2}} \left[ \x\x^{\top} + I \frac{\phi(\left\Vert \x \right\Vert_{2})\left\Vert \x \right\Vert_{2}}{\phi'(\left\Vert \x \right\Vert_{2})}\right] \text{, and,}\\
		\lv \nabla T_{\t}(\x) \rv  &= \left(\frac{\phi'(\left\Vert \x \right\Vert_{2})}{\left\Vert \x \right\Vert_{2}}\right)^n\lv \x\x^{\top} + I \frac{\phi(\left\Vert \x \right\Vert_{2})\left\Vert \x \right\Vert_{2}}{\phi'(\left\Vert \x \right\Vert_{2})}\rv.
		\end{align*}
		By Sylvester's determinant theorem we have
		\begin{align*}
		\lv \x\x^{\top} + I \frac{\phi(\left\Vert \x \right\Vert_{2})\left\Vert \x \right\Vert_{2}}{\phi'(\left\Vert \x \right\Vert_{2})} \rv &= \left(1 + \frac{\phi'(\left\Vert \x \right\Vert_{2})}{\phi(\left\Vert \x \right\Vert_{2})\left\Vert \x \right\Vert_{2}}\left\Vert \x \right\Vert_{2}^2 \right) \left(\frac{\phi(\left\Vert \x \right\Vert_{2})\left\Vert \x \right\Vert_{2}}{\phi'(\left\Vert \x \right\Vert_{2})}\right)^n\\
		\lv \nabla T_{\t}(\x) \rv &= \phi(\xn)^{n-1} \left(\phi(\xn) + \phi'(\xn)\xn \right)
		\end{align*}
		and since $\alpha(r) = \phi(r)r$ and $\alpha'(r) = \phi(r) + \phi'(r)r$, we have $\lv \nabla T_{\t}(\x) \rv = \phi(\xn)^{n-1}\alpha'(\xn)$.
	\end{proof}
\end{proposition}

For the learning task, since $\left\vert \nabla T_{\t}(\x) \right\vert = \phi_{n,\theta}(\xn)^{n-1}\alpha_{n,\theta}'(\xn)$ and  $T_\t^{-1}(\y) = \psi_{n,\theta}(\yn)\y  = \frac{\alpha^{-1}_{n,\theta}(\yn) }{\yn}\y$, we have that the complexity term is given by $$\log \lvert  \nabla S_{\t}(\y) \rvert = (n-1)\log(\alpha_{n,\theta}^{-1}(\yn)) -\log\left(\alpha_{n,\theta}'(\alpha_{n,\theta}^{-1}(\yn))\right).$$

\subsubsection{Inference on elliptical transport}
Since the reference distribution $\eta_\t$ is spherical, then $\eta_{\t}(\x) = \beta_n(\x^\top \x)$ for some positive function $\beta_n$. The transported distribution is also spherical with density $\pi_{\t}(\y)=h_{n}(\y^\top\y) := \beta_n(\psi_{n,\theta}^2(\yn)\y^\top\y)\psi_{n,\theta}(\yn)^{(n-1)}(\alpha_{n,\theta}^{-1})'(\yn)$. 

Given observations $(\t,\y)$, for inference on new inputs $\tto$ we have that the posterior distribution is also a spherical distribution, with density given by $\pi_{\tto|\t}(\yo|\y)=\frac{h_{n+\no}(\yo^\top\yo + \yn^2)}{h_{n}(\yn^2)}$. 

Since $\xo \sim \eta_{\tto}$ is spherical then $\frac{\xo}{\xon} \stackrel{d}{=} U^{(\no)}$, so if $\beta \sim p(\yon|\yn) $ is independent of $\frac{\xo}{\xon}$ then we have 
$$\yo|\y \stackrel{d}{=} \frac{ \beta }{\xon} \xo,$$
where $\beta$ is the positive r.v. of the norm of $\yo|\y$, that has density $$p(\yon|\yn) = \frac{2\pi^{\no/2}}{\Gamma(\no/2)}\yon^{\no-1}\frac{h_{n+\no}(\yon^2 + \yn^2) }{h_{2,n}(\yn^2)},$$

where $h_{2,n}$ is the marginal distribution of $\y$ from $(\y,\yo)$. We can generate samples efficiently: sampling $\xo$ is straightforward from $\eta$, and $\beta$ is an independent one dimensional positive random variable with explicit density. Note that $h_{2,n}(\yn^2)$ is the normalisation constant, so we can avoid its computation via MCMC methods like slice sampling or emcee sampling \cite{brooks2011handbook,neal2003slice,foreman2013emcee}.

\subsubsection{Student-t case}

The approach above includes the special case of the Student-t\footnote{The Student-t distribution, and Gaussian as its limit, is the unique elliptical distribution with positive density over all $\R^n$ that is closed under conditioning \cite{stoeber2013simplified}.} process as follows: Consider $R_\theta \sim \sqrt{\Gamma^{-1}(\frac{\theta}{2}, \frac{\theta}{2})}$ with $\Gamma^{-1}$ the inverse-gamma. Then $R_{n,\theta}:=R_n R_{\theta} \sim \sqrt{nF_{n,\theta}}$, where $F_{n,\theta}$ denote the Fisher–Snedecor distribution, and we have that $\pi_\t = \T_{n}(\theta, 0 , I_n)$ is a uncorrelated Student-t distribution with $\theta>2$ degrees of freedom. Given observations $\y$, the distribution has closed-form posteriors: $R_\theta|\y \sim \sqrt{\Gamma^{-1}(\frac{\theta+n}{2}, \frac{\theta+\yn^2}{2})}$ and $R_{\no,\theta}|\y  \sim \sqrt{\frac{\no(\theta+\yn^2)}{\theta+n} F_{\no,\theta+n}}$.  Also, for a bivariate Student-t distribution with correlation $\rho$ and degrees of freedom $\theta$, its copula has coefficients of tail dependence given by $\lambda_{u} = \lambda_{l} = 2t_{\theta +1}\left(-\frac{\sqrt{\theta +1}\sqrt{1-\rho}}{\sqrt{1+\rho}} \right) > 0$, strictly heavier that the Gaussian case.

\begin{figure}[h]
	\centering
	\includegraphics[width=0.9\textwidth]{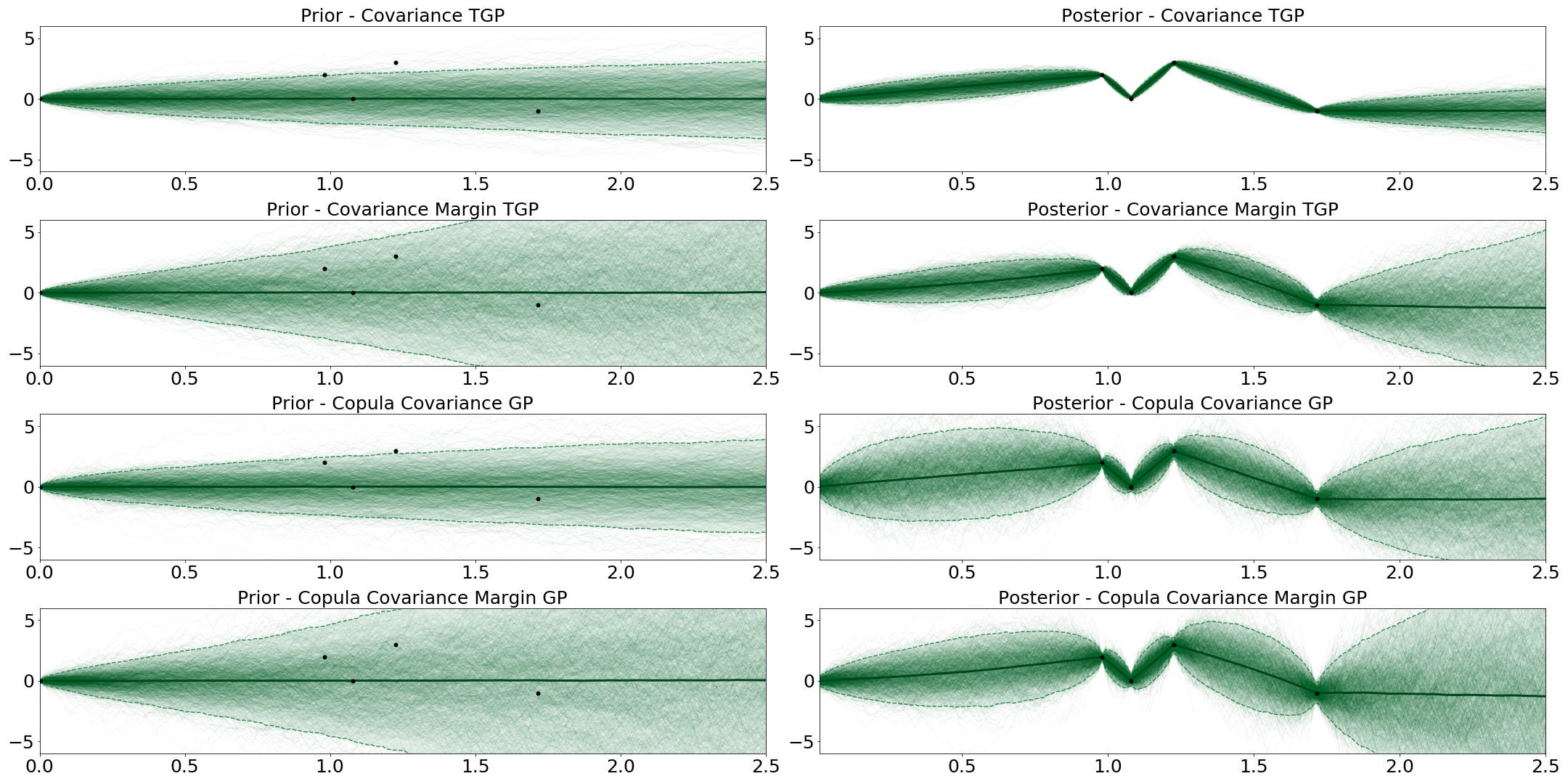}
	\caption{Samples from 4 TGP: the first and second examples have Gaussian copula, while third and fourth examples have Student-t copula.}
	\label{fig_tgp1}
\end{figure}

As an illustrative example, in Fig. \ref{fig_tgp1} we can see the mean (solid line), the 95\textbf{\%} confidence interval (dashed line) and 1000 samples (blurred lines) from 4 TGPs. All of them use a Brownian kernel $k(t,s) =min(t,s)$ for covariance transport, beside the second and fourth have an affine margin transport and the third and fourth have a Student-t elliptical transport. On the left column we plot the priors and on the right column we plot the posterior. The given observations are denoted with black dots. In this example we can see the difference between the Gaussian and Student-t copulas, although the priors look similar, the posteriors are quite different, where the Student-t copulas have more mass at the extrema. 

\subsection{Archimedean processes}

From a Gaussian reference, the previous transport allows the generation of any elliptical copula. However, our approach is more general, and it is possible to obtain non-elliptical copulas, specifically the so-called Archimedean copulas. 
\begin{definition}
	A copula $C(\u)$ is called Archimedean if it can be written in the form $C(\u) = \psi\left(\sum_{i=1}^{n}\psi^{-1}(u_i)\right)$ where $\psi:\R^{+} \to [0,1]$ is continuous, with $\psi(0)=1$, $\psi(\infty)=0$ and its generalized inverse $\psi^{-1}(x)=\inf\{u:\psi(u)\le x\}$.
\end{definition}
Archimedean copulas have explicit form for tail dependency: $\lambda_{l} = 2\lim\limits_{x\to0^+}\frac{\psi'(x)-\psi'(2x)}{\psi'(x)}$ and $\lambda_{u} = 2\lim\limits_{x\to \infty}\frac{\psi'(2x)}{\psi'(x)}$. 

For example, if we consider the generator $\psi(u) = \exp(-u)$ then their Archimedean copula coincides with the independence copula $C(\u) = \prod_{i=1}^{n}u_i$ and $\lambda_{l}=\lambda_{u}=0$. Some Archimedean copulas, like the independent one, can be extended as stochastic processes, which are characterised by the following proposition.

\begin{proposition}
	Let $\psi:\R^{+} \to [0,1]$ completely monotone, i.e. $\psi \in \mathcal{C}^{\infty}(\R^{+}, [0,1])$ and $(-1)^k \psi^{(k)}(x) \ge 0$ for $ k \ge 1$. Then there exists a stochastic process where there finite-dimensional laws are $C_n(\u) = \psi\left(\sum_{i=1}^{n}\psi^{-1}(u_i)\right)$.
	\begin{proof} 
		By Kimberling’s Theorem\cite{matthias2017simulating} $\psi$ generates an Archimedean copula in any dimension iff $\psi$ is completely monotone. Note that Archimedean copulas are exchangeable, i.e. for any $n$-permutation $\tau$ we have that $\u \stackrel{d}{=} \tau(\u)$, so in particular they are consistent under permutation, so we have that $F_{\eta_{\tau(\t)}}(\tau(\u)) = C_n(\tau(\u)) = C_n(\u) = F_{\eta_{\t}}(\u)$. The consistency under marginalisation is straightforward since $C_{n+1}(\u,1) = \psi\left(\sum_{i=1}^{n}\psi^{-1}(u_i) + \psi^{-1}(1)\right)=C_{n}(\u)$, and we conclude.
	\end{proof}
\end{proposition}

Any Archimedean copula process has a completely monotone generator $\psi$ associated that, by Bernstein’s Theorem\cite{matthias2017simulating}, is the Laplace transform \footnote{The Laplace transform of a random variable $Z>0$ is defined as $\L(Z)(s) = \E(\exp(-sZ))=\int_{0}^{\infty}e^{-sz}dF_Z(z)$ for $s \in [0,\infty]$. } of a positive distribution $F$, i.e. $\psi = \L[F]$ and $F = \L^{-1}[\psi]$. The following proposition shows the relation between Archimedean copulas and simplicial contoured distributions \cite{ghaffari2018multivariate, mcneil2009multivariate}..
\begin{proposition}
	\label{prop:l1_copula}
	Let $S_n \sim \Gamma(n,1)$, $W$ a real positive r.v. and $U^{[n]}$ a uniform r.v. on the unit simplex in $\R^{n}$ (i.e. $\lV U^{[n]} \rV_1 = 1$), where $S_n$, $W$ and $U^{[n]}$ are independent. Then $\x = (S_n/W)U^{[n]}$ follows a simplicial contoured distribution with an Archimedean survival copula generated by $\psi = \L[F_W]$, and each $x_i$ has marginal distribution $F_{x_i}(x)=1-\psi(x)$.
	\begin{proof}
		We have that $S_nU^{[n]} \stackrel{d}{=} (E_1,...,E_n)$ where $E_i \sim Exp(1)$ are independent. By Marshall and Olkin algorithm \cite{matthias2017simulating}, if $W \sim \L^{-1}[\psi]$ then $\v \sim C(\v) =  \psi\left(\sum_{i=1}^{n}\psi^{-1}(v_i)\right)$ where $v_i=\psi(x_i)$. Since the transport from $\x$ to $\v$ is diagonal, they share the same copula, so $\x$ also has copula $C(\v)$. Finally, since $\psi(x_i) = v_i \stackrel{d}{=} 1-v_i \sim \U[0,1]$ then $1-\psi(x_i)$ is the marginal distribution of each $x_i$ for $i=1,...,n$.
	\end{proof}
\end{proposition}

Simplicial distributions $\x \stackrel{d}{=} RU^{[n]}$, also know as $\ell_1$-norm symmetric distributions, satisfy $\xs = \sum_{i=1}^{n}x_i \stackrel{d}{=} R$ and $\frac{\x}{\xs} \stackrel{d}{=}  U^{[n]}$. If $R$ has density $p_R$ then $\x$ has density $p_{\x}(\x) = \Gamma(n) \xs^{1-n}p_R(\xs)$. For example, if the independence copula has generator $\psi(x) = \exp(-x)$ then $W$ is degenerate on $1$, so $R \stackrel{d}{=} S_n/W \sim \Gamma(n,1)$ and marginals distribute as $x_i \sim Exp(1)$. In another example, if $W \sim \Gamma(\frac{1}{\theta}, 1)$ then $\psi_\theta(s) = (1+s)^{-1/\theta}$ and $C(\u) = (\sum_{i=1}^{n}u_i^{-\theta}-n+1)^{-1/\theta}$, the so-called Clayton copula. We have that $R\stackrel{d}{=}S_n/W \sim \theta nF(2n,2/\theta)$ and marginals distribute as $F(x_i) = 1-(1+x_i)^{-1/\theta}$, a shifted Pareto distribution.

\subsubsection{Archimedean transport}

Note the similitude between spherical and simplicial distributions, changing the role of the $\ell_2$-norm by the $\ell_1$-norm. If $\y \stackrel{d}{=} S U^{[n]}$ for another real non-negative r.v. $S \in \R^{+}$, then the radial map $T^\alpha(\x) = \frac{F_{S}^{-1}(F_{R}(\xs))}{\xs}\x  \stackrel{d}{=} \frac{S}{R}\x\stackrel{d}{=} S U^{[n]}\stackrel{d}{=} \y$ is a transport map from $\x$ to $\y$. The next proposition shows how to transport a normal distribution into a simplicial distribution.

\begin{proposition}
	Let $\x \sim \N_n(0,I_n)$. Denote $\Phi$ the distribution function of standard normal and consider the marginal transport $T^h$ defined by $h(t,x) = -\log \Phi(x)$, i.e. $T^h(\x)_i = -\log(\Phi(x_i))$.  Given $S_n \stackrel{d}{=} R_n/W$ for a positive r.v. $W$ independent of $R_n \sim \Gamma(n,1)$, then the Archimedean transport $T^\alpha_n(\y) = \phi(\ys)\y = \frac{F_{S_n}^{-1}(F_{R_n}(\ys))}{\ys}\y$ satisfies that $T^\alpha_n \circ T^h(\x)$ has an Archimedean copula with generator $\psi = \L^{-1}(W)$.
	\begin{proof}
		If $x_i \sim \N(0,1)$ then $y_i = -\log(\Phi(x_i)) \sim Exp(1)$, so the sum satisfies that $\ys = \sum_{i=1}^{n}y_i  \sim \Gamma(n,1)$ so $\ys \stackrel{d}{=} R_n$. It is know that $\left(\frac{y_1}{\ys},...,\frac{y_n}{\ys}\right) \stackrel{d}{=}  U^{[n]}$ is independent from $\ys$, so $T^h(\x) = \y = \ys \frac{\y}{\ys} \stackrel{d}{=} R_n U^{[n]}$. As $T^\alpha_{n}$ is a radial transport, then $T^\alpha_n \circ T^h$ transports $\x$ into a simplicial distribution, and by the prop. \ref{prop:l1_copula}, we conclude.
	\end{proof}
\end{proposition}

The last proposition implies that the transport $T = \{T_{\t} | \t \in \T^n, n \in \mathbb{N}\}$, where $T_{\t}(\x) = T^\alpha_n \circ T^h(\x)$, is an $f$-transport with $f \sim \GP(0,\delta(t,\bar t))$, where the transport process $g := T(f)$ has a finite-dimensional Archimedean copula.

\subsubsection{Learning an Archimedean transport}
As the marginal transport was studied previously, we only need the \emph{model complexity penalty} for this radial map.

\begin{proposition}
	Given the map $T(\y) = \phi(\ys)\y = \frac{F_{S}^{-1}(F_{R}(\ys))}{\ys}\y$, then $\lv \nabla T_{\t}(\x) \rv = \phi(\xs)^{n-1}\alpha'(\xs)$.
	\begin{proof} Note that
		\begin{align*}
		\frac{\partial T_{\t}(\x)_i}{\partial x_i} &= \phi(\xs) + \phi'(\xs)x_i,\\
		\frac{\partial T_{\t}(\x)_i}{\partial x_j} &= \phi'(\xs)x_i, \text{if } i\neq j,\\
		\nabla T_{\t}(\x) &= \phi(\xs)I + \phi'(\xs) \x\mathbf{1}^\top =\phi'(\xs)\left[\frac{\phi(\xs)}{\phi'(\xs)}I + \x\mathbf{1}^\top\right].
		\end{align*}
		By Sylvester's determinant theorem we have
		\begin{align*}
		\lv \nabla T_{\t}(\x) \rv  &= \phi'(\xs)^n\left(\frac{\phi(\xs)}{\phi'(\xs)}\right)^n\left(1 +  \mathbf{1}^\top \left(\frac{\phi'(\xs)}{\phi(\xs)}I\right) \x\right),\\
		&= \phi(\xs)^{n-1}\left(\phi(\xs) +\phi'(\xs)\xs\right),\\
		&= \phi(\xs)^{n-1}\alpha'(\xs).
		\end{align*}
		thus concluding the proposed.
	\end{proof}
\end{proposition}

With the above result, we have that the \emph{model complexity penalty} is given by
\begin{align*}
\log \lv \nabla S_{\t}(\y) \rv &= -\log \lv \nabla T_{\t}(S_{\t}(\y)) \rv, \\
&= -(n-1)\log \left(\frac{\yn}{\alpha^{-1}(\yn)}\right)-\log \left( \alpha'(\alpha^{-1}(\yn)) \right),\\
&= -(n-1)\log \left(\frac{\yn}{\alpha^{-1}(\yn)}\right)+\log \left( \alpha^{-1}(\yn)' \right).
\end{align*}

\subsubsection{Inference with Archimedean transport}

For an Archimedean copula, the conditional distribution given $k$ observations $o_1,...,o_k$ is given by $C(\u|o_1,....,o_k)=\frac{\psi^{(k)}\left(\sum_{i=1}^{n}\psi^{-1}(u_i)+a\right)}{\psi^{(k)}(a)}$ where $a = \sum_{j=1}^{k}\psi^{-1}(o_j)$ and $\psi^{(k)}$ is the $k$-th derivative of the generator $\psi$. We can then use methods for sampling the conditional Archimedean $\u$,to then apply the diagonal push-forward via $F^{-1}(u_i)$ where $F(x) = 1-\psi(x)$.


\section{Deep Transport Process}
\label{sec:computationtgp}
Both the generality and the feasible calculation of the presented transport-based approach to non-parametric regression motivate us to define complex models inspired on recent advances from the deep learning community. Via the composition of elementary transports (or \emph{layers}) we can generate more expressive (or \emph{deep}) transports. In this section, we will explain how to build such an architecture, describe the properties that are inherited through the composition, to finally propose families of transports that can be composed together and study their properties in the regression problem.

\subsection{Consistent deep transport process}

In this paper we introduce four types of transports, that can be seen as elementary \emph{layers} for regression models. Our approach starts from a Gaussian white noise reference $f \sim \eta$, since it is a well-know process with explicit density and efficient sampling methods. The first layer determines the \emph{copula} of the induced process, that can be elliptical or Archimedian via elliptical or Archimedian transports. In the elliptical case, it is possible to compose it with a covariance transport in order to determine the correlation on the induced stochastic process. Finally, in any case, we can compose any number of marginal transports to define an expressive marginal distribution over the induced stochastic process, as it is shown in the previous work \cite{cwgp}. As we saw in the previous sections, these compositions are consistent and expressive enough to include GPs, warped GPs, Student-t processes, Archimedean processes, elliptical processes, and those that we could call \emph{warped Archimedean processes} and \emph{warped elliptical processes}.

\subsection{Learning deep transport process}
Assume $T\#\eta = \pi$, where $T$ is the composition of $k$ transports, i.e. $T = T^{(k)} \circ ... \circ T^{(1)}$. Denote $\eta^{(0)}=\eta$ and assume that each $\eta^{(j)}=T^{(j)}\#\eta^{(j-1)}$ is a transport process with finite-dimensional transports $\{T_{\t}^{(j)}\}_{j=1}^k$. Note that $\eta^{(k)} = T\#\eta = \pi$, where $T_{\t} = T_{\t}^{(k)} \circ ... \circ T_{\t}^{(1)}$ are finite-dimensional transports with $S_{\t} = S_{\t}^{(1)} \circ ... \circ S_{\t}^{(k)}$. As a consequence, the composition of transport processes is a transport process. Consequently, the NLL can be calculated as
\begin{align}
\label{eq:TGP_NLL2}
-\log \pi_{\t}(\y|\theta) = -\log \eta_{\t}(S_{\t}(\y)) - \sum\nolimits_{j=1}^k \log |\nabla S_{\t}^{(j)}(S_{\t}^{[(j+1):k]}(\y))|,
\end{align}
where $S_{\t}^{[j:k]}(\y) = S_{\t}^{(j)} \circ ... \circ S_{\t}^{(k)} (\y)$, with the convention $S_{\t}^{[(k+1):k]}(\y) = \y$. The formula above is based on calculating each $F_{\t}^{(j)}(\z)= \log |\nabla S_{\t}^{(j)}(\z)|$, which can be computed alternatively as $F_{\t}^{(j)}(\z)=- \log |\nabla T_{\t}^{(j)}(S_{\t}^{(j)}(\z))|$, or, for the triangular case, as $F_{\t}^{(j)}(\z)=\sum_{i} \log \frac{\partial (S_{\t})_i}{\partial y_i}(\z)$. The following algorithm computes the NLL, subject to being able to evaluate each function $F_{\t}^{(j)}$ and $S_{\t}^{(j)}$.

\begin{algorithm}[!h]
	\caption{Calculate the NLL of a deep transport process}
	\label{algo1}
	\begin{algorithmic} 
		\REQUIRE Data $(\t,\y )$, inverse transports $T^{-1}_{\t}(\z) = S_{\t}^{(1)} \circ ... \circ S_{\t}^{(k)}(\z)$ and $F_{\t}^{(j)}(\z)=\log |\nabla S_{\t}^{(j)}(\z)|$.
		\ENSURE  $\mathcal{L} = -\log \pi_{\t}(\y|\theta)$
		\STATE $\z \gets \y$, $\mathcal{L} \gets 0$
		\FOR{$ j \in k,...,1 $}
		\STATE $\mathcal{L} \gets \mathcal{L} - F_{\t}^{(j)}(\z)$
		\STATE $\z \gets S_{\t}^{(j)}(\z)$
		\ENDFOR
		\STATE $\mathcal{L} \gets \mathcal{L} - \log \eta_{\t}(\z)$
		\RETURN $\mathcal{L}$
	\end{algorithmic}
\end{algorithm}

\begin{remark}
	Algorithm \ref{algo1} is based in applying the chain rule and the inverse function theorem over the composited inverse  $S_{\t} = S_{\t}^{(1)} \circ ... \circ S_{\t}^{(k)}$, so
	\begin{align}
	\nabla S_{\t}(\y) &= \nabla S_{\t}^{(1)}( S_{\t}^{(2)} \circ ... \circ S_{\t}^{(k)}) \nabla S_{\t}^{(2)}( S_{\t}^{(3)} \circ ... \circ S_{\t}^{(k)}) .... \nabla S_{\t}^{(k-1)}(S_{\t}^{(k)}(\y))  \nabla S_{\t}^{(k)}(\y),\\
	&= \nabla T_{\t}^{(1)}( S_{\t}^{(1)} \circ ... \circ S_{\t}^{(k)})^{-1} \nabla T_{\t}^{(2)}( S_{\t}^{(2)} \circ ... \circ S_{\t}^{(k)})^{-1} .... \nabla T_{\t}^{(k)}(S_{\t}^{(k)}(\y))^{-1}.
	\end{align}
\end{remark}

Algorithm \ref{algo1} is computationally efficient in terms of minimal use of memory (even the variable $\z$ can use the same memory as $\y$), and it can be executed in the shortest possible time by calling each function $F_{\t}^{(j)}$ and $S_{\t}^{(j)}$ only once. By implementing the calculations of NLL in any modern tensor framework, such as PyTorch, it is possible to apply automatic differentiation \cite{paszke2017automatic} to calculating the derivative of NLL with respect to parameters. Additionally, this algorithm is parallelizable in $\theta$, thus allowing an efficient evaluation of NLL for multiple values for $\theta$ simultaneously in architectures such as GPUs. This is a desired property for derivative-free optimization methods such as particle swarm optimization \cite{kennedy2010particle}, or MCMC ensemble samplers \cite{goodman2010ensemble}. In stochastic gradient descent methods \cite{bottou2010large}, given that in each step we use a subsampling from the data, we can take advantage of the GPU-based architectures running in parallel multiple executions, in order to better navigate the space of models.


\subsection{Inference deep transport process}

As the composition operation preserves triangularity, we assume $T^{(j)}$ are triangular for $j > l$, in addition to being able to calculate the posterior of $\eta^{(l)}$, i.e. compute $\eta_{\tto|\t}^{(l)}(\cdot|\x)$ for any input $\tto$. Without loss of generality, it can be assumed that $ l = 1 $, since it is possible to collapse by composition the $l$ transports in only one. The following algorithm generates samples from the posterior distribution $\pi_{\tto|\t}(\yo|\y)$ under the above assumptions.

\begin{algorithm}[!h]
	\caption{Generate samples from the posterior}
	\label{algo2}
	\begin{algorithmic} 
		\REQUIRE Observations $\y \sim \pi_{\t}$, new inputs $\tto \in \I^d, d \in \mathbb{N}$, number of samples $N \in \mathbb{N}$.
		\ENSURE  $\yo_i \sim \pi_{\tto|\t}(\yo|\y)$ for $i=1,...,N$
		\STATE $\x \gets S_{\t}^{[l+1:k]}(\y)$
		\STATE $R(\cdot) \gets P_{\tto} \circ T_{\t,\tto}^{[l+1:k]}(\x, \cdot)$
		\FOR{$ i \in 1,...,N $}
		\STATE $\xo_i \sim \eta_{\tto|\t}^{(l)}(\cdot|\x)$
		\STATE $\yo_i \leftarrow R(\xo_i)$
		\ENDFOR
		\RETURN $\{\yo_1,...,\yo_N\}$
	\end{algorithmic}
\end{algorithm}

Algorithm \ref{algo2} is parallelisable in $N$, since the function $R(\cdot)$ is the same for all samples, and thus allows us to obtain multiple samples simultaneously in an efficient manner. This can be used in turn to calculate moments, quantiles or other statistics in an empirical way through Monte Carlo. 

\subsection{Noise layer}

Under the presence of noisy observations, following the same rationale as GPs, warped GPs \cite{snelson2004warped} and Student-t processes \cite{shah2014student}, we consider that the covariance transport has a special behavior. Let $k(t,s) = r(t,s) + \sigma_0\delta_{t,s}$, where $\delta$ is Kronecker delta, $\sigma_0$ is the parameter that controls the intensity of noise and $r(t,s)$ is the noise-free covariance function. We consider that the observations have uncorrelated noise. While for training we use $k(t,s)$ in the formula for NLL, in inference we use $k(t,s)$ on the backward-step (i.e. for the inverse map $\x = T_{\t}^{-1}(\y)$), and on the forward-step (i.e. for push-forward the reference distribution) we use $r(t,s)$, instead of $k(t,s)$, to perform a free-noise prediction.

\subsection{Sparse layer}
While marginal and copula transports can be evaluated efficiently without needing training data, the covariance transports needs all the data $\y$ to performance inference. The computational complexity of evaluation is $\O(n^2)$ in memory and $\O(n^3)$ in time, where $n=|\y|$. Sparse approximations are widely used to solve this issue on GPs \cite{quinonero2005unifying, snelson2006sparse, titsias2009variational}, and it is natural to define a \emph{sparse} transport as $T_{\tto}(\u) = \Sigma_{\tto\s}\Sigma_{\s\s}^{-1}\z + \chol(\Sigma_{\tto\tto} - \Sigma_{\tto\s}\Sigma_{\s\s}^{-1}\Sigma_{\tto\s})\u$, where $(\s,\z)$ are trainable pseudo-data with $|\s| = m < n$. The training of pseudo-data follows the same ideas that sparse GPs, like SoD and SoR approximations \cite{quinonero2005unifying}, where the computational cost drops to $\O(nm)$ in space and $\O(nm^2)$ in time.

\section{Experimental validation}
\label{sec:experimentstgp}

We validate our approach with three real-world time series, described as follows:
\begin{enumerate}
	\item \textbf{Sunspots Data}: The Sunspot time series \cite{sunspots} corresponds to the yearly number of sunspots between 1700 and 2008, resulting in 309 data points, one per year. These measures are positive and semi-periodic, with a cycle period of around 11-years.
	\item \textbf{Heart Data}: This is a heart-rate time series from the MIT-BIH Database (ecg.mit.edu) \cite{glass2012theory}. This series contains 1800 evenly-spaced positive measurements of instantaneous heart rate (in units of beats per minute) from a single subject, happening at 0.5 second intervals, and showing a semi-periodic pattern. For performance issues, we take a subsample of 450 measures at 2.0 seconds intervals.
	\item \textbf{Economic Data}: This time series corresponds to the quarterly average \emph{3-Month Treasury Bill: Secondary Market Rate} \cite{tb3ms} between the first quarter of 1959 and the third quarter of 2009, that is, 203 observations, one per quarter. We know beforehand that this macroeconomic signal is the price of U.S. government risk-free bonds, which cannot take negative values and can have large positive deviations.
\end{enumerate}

Due to the semi-periodic nature of the time series, we consider a noisy spectral mixture with two components kernel $k_{SM}$ \cite{wilson2013gaussian} for the covariance transport. Since the time series are positive, we use a shifted Box-Cox warping $\phi_{BC}$ \cite{rios2018learning} for marginal transport. We compare two models: a warped GP, with $k_{SM}$ kernel and $\phi_{BC}$ warping; and a TGP with a Student-t copula transport, besides the above-described covariance and marginal transports. 

\begin{figure}
	\centering
	\includegraphics[width=0.7\textwidth]{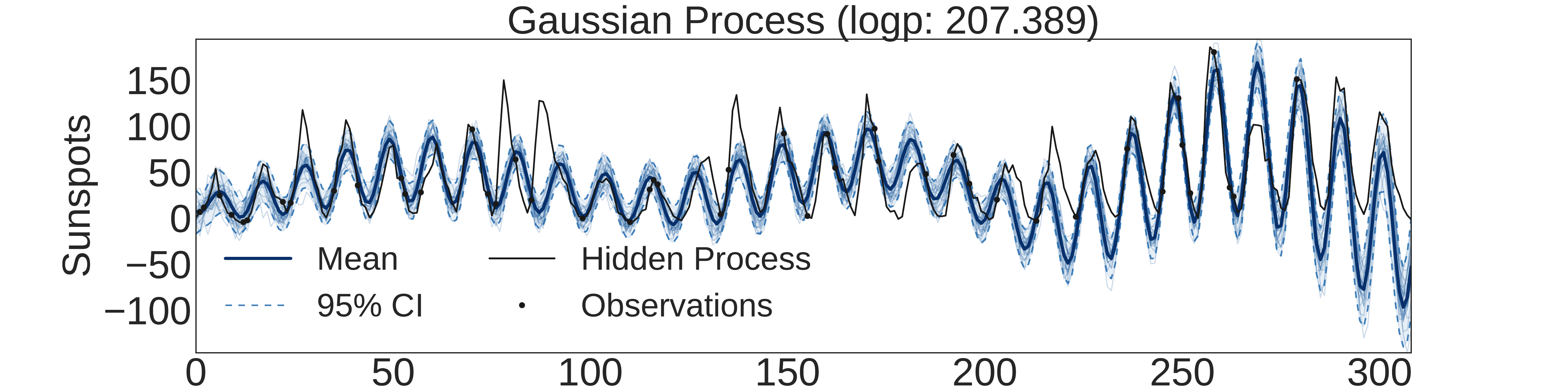}
	\includegraphics[width=0.7\textwidth]{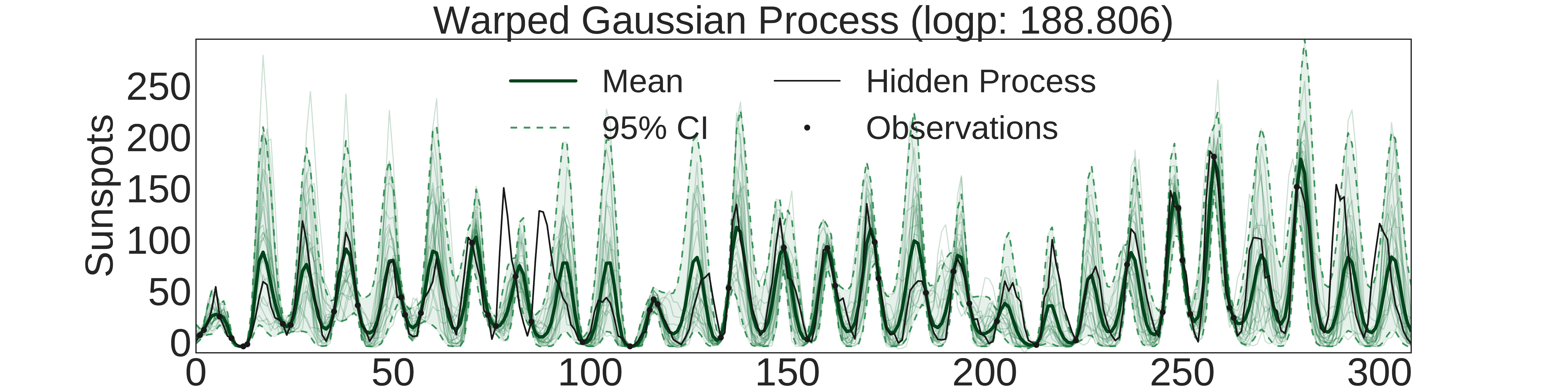}
	\includegraphics[width=0.7\textwidth]{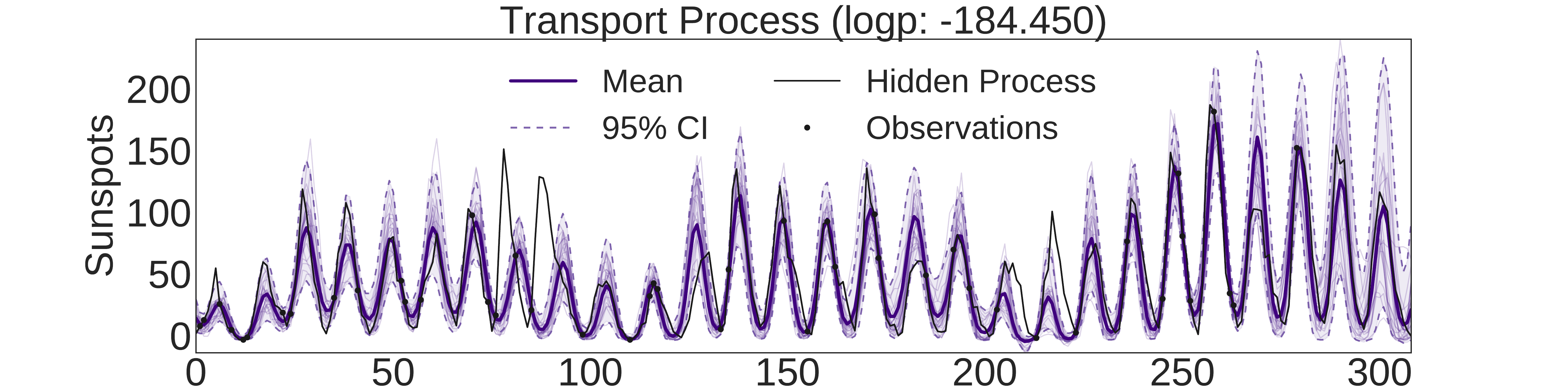}
	\caption{GP (blue), WGP (green) and TGP (purple) over Sunspots data.}
	\label{fig_sunspots}
\end{figure}

We leave the standard GPs out of the experiment since the assumption of Gaussianity violates the nature of the datasets, having a lower predictive power than the WGP, as shown in \cite{rios2018learning, riostobar2019cwgp}. To illustrate this fact, in Fig. \ref{fig_sunspots} we show the posterior of three trained models: GP in blue, WGP in green and TGP in purple. We plot the observations (black dots), the mean (solid line), the 95\textbf{\%} confidence interval (dashed line) and 25 samples (blurred lines). Notice how the GP fails to model the positivity and the correct amplitude of the phenomena.

\begin{table}
	\centering
	\tiny
	\begin{tabular}{l|ll|ll|ll}
		\toprule
		{} & \multicolumn{2}{l}{Sunspots} & \multicolumn{2}{l}{Heart} & \multicolumn{2}{l}{Economic} \\
		{} &                      WGP &                      TGP &                 WGP &                 TGP &                WGP &                TGP \\
		\midrule
		MAE &       25.266 $\pm$ 4.607 &       24.710 $\pm$ 4.271 &   2.965 $\pm$ 0.827 &   2.907 $\pm$ 0.715 &  1.132 $\pm$ 0.260 &  1.111 $\pm$ 0.215 \\
		EAE &       30.166 $\pm$ 4.374 &       29.649 $\pm$ 4.168 &   3.431 $\pm$ 0.732 &   3.388 $\pm$ 0.660 &  1.392 $\pm$ 0.235 &  1.380 $\pm$ 0.206 \\
		MSE &  1,306.253 $\pm$ 560.496 &  1,223.257 $\pm$ 421.385 &  16.405 $\pm$ 8.809 &  15.740 $\pm$ 7.619 &  3.002 $\pm$ 1.643 &  2.860 $\pm$ 1.311 \\
		ESE &  1,889.318 $\pm$ 633.325 &  1,796.989 $\pm$ 514.193 &  21.963 $\pm$ 8.524 &  21.554 $\pm$ 8.213 &  4.376 $\pm$ 1.725 &  4.272 $\pm$ 1.424 \\
		\bottomrule
	\end{tabular}
	\vspace{0.5em}
	\caption{WGP and TGP results over Sunspots, Heart and Economic datasets.}
	\label{tab:table_results} 
\end{table}

The experiment was implemented in a Python-based library named \emph{tpy: Transport processes in Python}\cite{tpy}, with a PyTorch backend for GPU-support and automatic differentiation \cite{paszke2017automatic}. The training was performed by minimising the NLL from eq.~\eqref{eq:TGP_NLL2}, via a stochastic mini-batches rprop method \cite{riedmiller1993direct}, to then end with non-stochastic iterations.

In each experiment, we randomly (uniformly) select 15\text{\%} of the data for training and the remaining 85\text{\%} for validation. Given the validation data points $\{y_i\}_{i=1}^n$, for each model we generate $S$ samples $\{y_i^{(k)}\}_{i=1}^n$ for $k=1,...,S$, and then we calculate four performance indices: the mean square error as $\text{MSE} = \frac{1}{n}\sum\limits_{i=1}^{n}\left(y_{i} - \frac{1}{S}\sum_{k=1}^{S}y_i^{(k)}\right)^2$, the mean absolute error as $\text{MAE} = \frac{1}{n}\sum\limits_{i=1}^{n}|y_{i} - \frac{1}{S}\sum_{k=1}^{S}y_i^{(k)}|$, the expected square error as $\text{ESE} =  \frac{1}{n}\sum\limits_{i=1}^{n}\frac{1}{S}\sum_{k=1}^{S}(y_{i} - y_{i}^{(k)})^2$, and the expected absolute error as $\text{EAE} = \frac{1}{n}\sum\limits_{i=1}^{n}\frac{1}{S}\sum_{k=1}^{S}|y_{i} - y_{i}^{(k)}|$. We repeat each experiment 100 times. The results for all of these experiments are summarized in Table 1, showing each mean and standard deviation. Consistently, the proposed TGP has better performance that the warped GP alternative, for each dataset and evaluation index.


\section{Conclusions}
\label{sec:conclusionstgp}

In this paper we have proposed a regression model from a unifying point of view with other approaches to literature, like GPs, warped GPs, Student-t processes and copula processes. We deliver the standard methods of training and inference. We hope to continue developing this work in the near future, heightening the relationship with deep learning and our methodologies, and expanding our work for multi-outputs and other types of data.

\section*{Acknowledgments}
We are grateful for the financial support from Conicyt \#AFB170001 Center for Mathematical Modeling and Conicyt-Pcha/DocNac/2016-21161789. We thank Felipe Tobar, Julio Backhoff and Joaqu\'in Fontbona for their valuable feedback and comments during the development of this work.

\bibliography{bibliography}

\begin{thebibliography}{10}

\bibitem{bottou2010large}
L{\'e}on Bottou.
\newblock Large-scale machine learning with stochastic gradient descent.
\newblock In {\em Proceedings of COMPSTAT'2010}, pages 177--186. Springer,
  2010.

\bibitem{brooks2011handbook}
Steve Brooks, Andrew Gelman, Galin Jones, and Xiao-Li Meng.
\newblock {\em Handbook of markov chain monte carlo}.
\newblock CRC press, 2011.

\bibitem{bui2016deep}
Thang Bui, Daniel Hern{\'a}ndez-Lobato, Jose Hernandez-Lobato, Yingzhen Li, and
  Richard Turner.
\newblock Deep gaussian processes for regression using approximate expectation
  propagation.
\newblock In {\em International Conference on Machine Learning}, pages
  1472--1481, 2016.

\bibitem{coles2001introduction}
Stuart Coles, Joanna Bawa, Lesley Trenner, and Pat Dorazio.
\newblock {\em An introduction to statistical modeling of extreme values},
  volume 208.
\newblock Springer, 2001.

\bibitem{cressie1990origins}
Noel Cressie.
\newblock The origins of kriging.
\newblock {\em Mathematical geology}, 22(3):239--252, 1990.

\bibitem{cuestaalbertos1993optimal}
Juan Cuesta-Albertos, L~Ruschendorf, and Araceli Tuero-Diaz.
\newblock Optimal coupling of multivariate distributions and stochastic
  processes.
\newblock {\em Journal of Multivariate Analysis}, 46(2):335--361, 1993.

\bibitem{damianou2015deep}
Andreas Damianou.
\newblock {\em Deep Gaussian processes and variational propagation of
  uncertainty}.
\newblock PhD thesis, University of Sheffield, 2015.

\bibitem{damianou2013deep}
Andreas Damianou and Neil Lawrence.
\newblock Deep gaussian processes.
\newblock In {\em Artificial Intelligence and Statistics}, pages 207--215,
  2013.

\bibitem{damianou2016variational}
Andreas~C Damianou, Michalis~K Titsias, and Neil~D Lawrence.
\newblock Variational inference for latent variables and uncertain inputs in
  gaussian processes.
\newblock {\em The Journal of Machine Learning Research}, 17(1):1425--1486,
  2016.

\bibitem{demarta2005t}
Stefano Demarta and Alexander~J McNeil.
\newblock The t copula and related copulas.
\newblock {\em International Statistical Review/Revue Internationale de
  Statistique}, pages 111--129, 2005.

\bibitem{demmel1997applied}
James~W Demmel.
\newblock {\em Applied numerical linear algebra}, volume~56.
\newblock Siam, 1997.

\bibitem{donnelly2010devil}
Catherine Donnelly and Paul Embrechts.
\newblock The devil is in the tails: actuarial mathematics and the subprime
  mortgage crisis.
\newblock {\em ASTIN Bulletin: The Journal of the IAA}, 40(1):1--33, 2010.

\bibitem{duvenaud2014avoiding}
David Duvenaud, Oren Rippel, Ryan Adams, and Zoubin Ghahramani.
\newblock Avoiding pathologies in very deep networks.
\newblock In {\em Artificial Intelligence and Statistics}, pages 202--210,
  2014.

\bibitem{tb3ms}
{Federal Reserve Bank of St. Louis}.
\newblock Federal reserve economic data, 2009.

\bibitem{foreman2013emcee}
Daniel Foreman-Mackey, David~W Hogg, Dustin Lang, and Jonathan Goodman.
\newblock emcee: the mcmc hammer.
\newblock {\em Publications of the Astronomical Society of the Pacific},
  125(925):306, 2013.

\bibitem{ghaffari2018multivariate}
Novin Ghaffari and Stephen Walker.
\newblock On multivariate optimal transportation.
\newblock {\em arXiv preprint arXiv:1801.03516}, 2018.

\bibitem{glass2012theory}
Leon Glass, Peter Hunter, and Andrew McCulloch.
\newblock {\em Theory of heart: biomechanics, biophysics, and nonlinear
  dynamics of cardiac function}.
\newblock Springer Science \& Business Media, 2012.

\bibitem{goodman2010ensemble}
Jonathan Goodman and Jonathan Weare.
\newblock Ensemble samplers with affine invariance.
\newblock {\em Communications in applied mathematics and computational
  science}, 5(1):65--80, 2010.

\bibitem{hogg1995introduction}
Robert~V. Hogg and Allen~T. Craig.
\newblock {\em Introduction to Mathematical Statistics}.
\newblock Upper Saddle River, New Jersey: Prentice Hall, fifth edition, 1995.

\bibitem{Sinharcsinh}
M.~Chris Jones and Arthur Pewsey.
\newblock Sinh-{A}rcsinh distributions.
\newblock {\em Biometrika}, 96(4):761, 2009.

\bibitem{kelker1970distribution}
Douglas Kelker.
\newblock Distribution theory of spherical distributions and a location-scale
  parameter generalization.
\newblock {\em Sankhy{\=a}: The Indian Journal of Statistics, Series A}, pages
  419--430, 1970.

\bibitem{kennedy2010particle}
James Kennedy.
\newblock Particle swarm optimization.
\newblock {\em Encyclopedia of machine learning}, pages 760--766, 2010.

\bibitem{kingma2013auto}
Diederik~P Kingma and Max Welling.
\newblock Auto-encoding variational bayes.
\newblock {\em arXiv preprint arXiv:1312.6114}, 2013.

\bibitem{krauth2016autogp}
Karl Krauth, Edwin~V Bonilla, Kurt Cutajar, and Maurizio Filippone.
\newblock Autogp: Exploring the capabilities and limitations of gaussian
  process models.
\newblock {\em arXiv preprint arXiv:1610.05392}, 2016.

\bibitem{lawrence2004gaussian}
Neil~D Lawrence.
\newblock Gaussian process latent variable models for visualisation of high
  dimensional data.
\newblock In {\em Advances in neural information processing systems}, pages
  329--336, 2004.

\bibitem{lazaro2012bayesian}
Miguel L{\'a}zaro-Gredilla.
\newblock Bayesian warped gaussian processes.
\newblock In {\em Advances in Neural Information Processing Systems}, pages
  1619--1627, 2012.

\bibitem{li2016review}
Ping Li and Songcan Chen.
\newblock A review on gaussian process latent variable models.
\newblock {\em CAAI Transactions on Intelligence Technology}, 1(4):366--376,
  2016.

\bibitem{matthias2017simulating}
Scherer Matthias and Mai Jan-frederik.
\newblock {\em Simulating copulas: stochastic models, sampling algorithms, and
  applications}, volume~6.
\newblock \# N/A, 2017.

\bibitem{mcneil2009multivariate}
Alexander~J McNeil, Johanna Ne$\check{s}$lehov{\'a}, et~al.
\newblock Multivariate archimedean copulas, d-monotone functions and
  $\ell1$-norm symmetric distributions.
\newblock {\em The Annals of Statistics}, 37(5B):3059--3097, 2009.

\bibitem{neal2003slice}
Radford~M Neal et~al.
\newblock Slice sampling.
\newblock {\em The annals of statistics}, 31(3):705--767, 2003.

\bibitem{owen1983class}
Joel Owen and Ramon Rabinovitch.
\newblock On the class of elliptical distributions and their applications to
  the theory of portfolio choice.
\newblock {\em The Journal of Finance}, 38(3):745--752, 1983.

\bibitem{paszke2017automatic}
Adam Paszke, Sam Gross, Soumith Chintala, Gregory Chanan, Edward Yang, Zachary
  DeVito, Zeming Lin, Alban Desmaison, Luca Antiga, and Adam Lerer.
\newblock Automatic differentiation in pytorch.
\newblock 2017.

\bibitem{petersen2008matrix}
Kaare~Brandt Petersen, Michael~Syskind Pedersen, et~al.
\newblock The matrix cookbook.
\newblock {\em Technical University of Denmark}, 7(15):510, 2008.

\bibitem{quinonero2005unifying}
Joaquin Qui{\~n}onero-Candela and Carl~Edward Rasmussen.
\newblock A unifying view of sparse approximate gaussian process regression.
\newblock {\em Journal of Machine Learning Research}, 6(Dec):1939--1959, 2005.

\bibitem{rasmussen06}
C.~E. Rasmussen and C.~K.~I. Williams.
\newblock {\em Gaussian Processes for Machine Learning}.
\newblock MIT, 2006.

\bibitem{riedmiller1993direct}
Martin Riedmiller and Heinrich Braun.
\newblock A direct adaptive method for faster backpropagation learning: The
  rprop algorithm.
\newblock In {\em Proceedings of the IEEE international conference on neural
  networks}, volume 1993, pages 586--591. San Francisco, 1993.

\bibitem{tpy}
Gonzalo Rios.
\newblock Tpy: Transport processes in python, github.com/griosd/tpy, 2017.

\bibitem{rios2018learning}
Gonzalo Rios and Felipe Tobar.
\newblock Learning non-{G}aussian time series using the {B}ox-{C}ox {G}aussian
  process.
\newblock In {\em 2018 International Joint Conference on Neural Networks
  (IJCNN)}, pages 1--8. IEEE, 2018.

\bibitem{riostobar2019cwgp}
Gonzalo Rios and Felipe Tobar.
\newblock Compositionally-warped {G}aussian processes.
\newblock {\em Neural Networks}, 118:235--246, 2019.

\bibitem{rohatgiintroduction}
VK~Rohatgi.
\newblock An introduction to probability theory and mathematical statistics.
  1976.

\bibitem{rubinstein2016simulation}
Reuven~Y Rubinstein and Dirk~P Kroese.
\newblock {\em Simulation and the Monte Carlo method}, volume~10.
\newblock John Wiley \& Sons, 2016.

\bibitem{rudin1964principles}
Walter Rudin et~al.
\newblock {\em Principles of mathematical analysis}, volume~3.
\newblock McGraw-hill New York, 1964.

\bibitem{salimbeni2017doubly}
Hugh Salimbeni and Marc Deisenroth.
\newblock Doubly stochastic variational inference for deep gaussian processes.
\newblock In {\em Advances in Neural Information Processing Systems}, pages
  4588--4599, 2017.

\bibitem{schmidt2005tail}
Rafael Schmidt.
\newblock Tail dependence.
\newblock In {\em Statistical Tools for Finance and Insurance}, pages 65--91.
  Springer, 2005.

\bibitem{shah2014student}
Amar Shah, Andrew~Gordon Wilson, and Zoubin Ghahramani.
\newblock Student-t processes as alternatives to {G}aussian processes.
\newblock In {\em AISTATS}, pages 877--885, 2014.

\bibitem{shahriari2015taking}
Bobak Shahriari, Kevin Swersky, Ziyu Wang, Ryan~P Adams, and Nando De~Freitas.
\newblock Taking the human out of the loop: A review of bayesian optimization.
\newblock {\em Proceedings of the IEEE}, 104(1):148--175, 2015.

\bibitem{shalizi2010almost}
Cosma~Rohilla Shalizi and Aryeh Kontorovich.
\newblock Almost none of the theory of stochastic processes.
\newblock {\em Lecture Notes}, 2010.

\bibitem{sunspots}
{SILSO World Data Center}.
\newblock {The International Sunspot Number}.
\newblock {\em International Sunspot Number Monthly Bulletin and online
  catalogue}, 1700-2008.

\bibitem{sklar1959fonctions}
M~Sklar.
\newblock Fonctions de repartition an dimensions et leurs marges.
\newblock {\em Publ. inst. statist. univ. Paris}, 8:229--231, 1959.

\bibitem{snelson2006sparse}
Edward Snelson and Zoubin Ghahramani.
\newblock Sparse gaussian processes using pseudo-inputs.
\newblock In {\em Advances in neural information processing systems}, pages
  1257--1264, 2006.

\bibitem{snelson2004warped}
Edward Snelson, Zoubin Ghahramani, and Carl~E Rasmussen.
\newblock Warped gaussian processes.
\newblock In S.~Thrun, L.~K. Saul, and P.~B. Sch\"{o}lkopf, editors, {\em
  Advances in neural information processing systems}, volume~16, pages
  337--344. MIT Press, 2004.

\bibitem{solin2015state}
Arno Solin and Simo S{\"a}rkk{\"a}.
\newblock State space methods for efficient inference in student-t process
  regression.
\newblock In {\em Artificial Intelligence and Statistics}, pages 885--893,
  2015.

\bibitem{stein2012interpolation}
Michael~L Stein.
\newblock {\em Interpolation of spatial data: some theory for kriging}.
\newblock Springer Science \& Business Media, 2012.

\bibitem{stoeber2013simplified}
Jakob Stoeber, Harry Joe, and Claudia Czado.
\newblock Simplified pair copula constructions--limitations and extensions.
\newblock {\em Journal of Multivariate Analysis}, 119:101--118, 2013.

\bibitem{tao2011introduction}
Terence Tao.
\newblock {\em An Introduction to Measure Theory}, volume 126.
\newblock American Mathematical Society, 2011.

\bibitem{titsias2009variational}
Michalis Titsias.
\newblock Variational learning of inducing variables in sparse {G}aussian
  processes.
\newblock In David van Dyk and Max Welling, editors, {\em Proc. of the
  International Conference on Artificial Intelligence and Statistics},
  volume~5, pages 567--574, 2009.

\bibitem{titsias2010bayesian}
Michalis Titsias and Neil~D Lawrence.
\newblock Bayesian gaussian process latent variable model.
\newblock In {\em Proceedings of the Thirteenth International Conference on
  Artificial Intelligence and Statistics}, pages 844--851, 2010.

\bibitem{wang2016sequential}
Yali Wang, Marcus Brubaker, Brahim Chaib-Draa, and Raquel Urtasun.
\newblock Sequential inference for deep gaussian process.
\newblock In {\em Artificial Intelligence and Statistics}, pages 694--703,
  2016.

\bibitem{wilson2013gaussian}
Andrew Wilson and Ryan Adams.
\newblock Gaussian process kernels for pattern discovery and extrapolation.
\newblock In {\em International Conference on Machine Learning}, pages
  1067--1075, 2013.

\bibitem{wilson2010copula}
Andrew Wilson and Zoubin Ghahramani.
\newblock Copula processes.
\newblock In J.~D. Lafferty, C.~K.~I. Williams, J.~Shawe-Taylor, R.~S. Zemel,
  and A.~Culotta, editors, {\em Advances in Neural Information Processing
  Systems 23}, pages 2460--2468. Curran Associates, Inc., 2010.

\bibitem{wilson2011gaussian}
Andrew~Gordon Wilson, David~A Knowles, and Zoubin Ghahramani.
\newblock Gaussian process regression networks.
\newblock {\em arXiv preprint arXiv:1110.4411}, 2011.

\end{thebibliography}
\bibliographystyle{plain}

\end{document}